\documentclass{article}
\usepackage{ijcai22}

\usepackage{times}
\usepackage{soul}
\usepackage{url}
\usepackage[hidelinks]{hyperref}
\usepackage[utf8]{inputenc}
\usepackage[small]{caption}
\usepackage{graphicx}
\usepackage{amsmath}
\usepackage{amsthm}
\usepackage{booktabs}
\usepackage{algorithm}
\usepackage{algorithmic}
\usepackage{amssymb,amsfonts}
\usepackage{booktabs}
\usepackage{multirow}
\usepackage{subfigure}

\newtheorem{theorem}{Theorem}

\newtheorem{remark}{Remark}
\newtheorem{corollary}{Corollary}

\newtheorem{assumption}{Assumption}
\newtheorem{definition}{Definition}

\title{Understanding the Generalization Performance of Spectral Clustering Algorithms}

\author{
Shaojie Li$^{1,2,\ast}$
\and
Sheng Ouyang$^{1,2,}$\thanks{Equal contribution.}\And
Yong Liu$^{1,2,}$\footnote{Corresponding author.}
\affiliations
$^1$Gaoling School of Artificial Intelligence, Renmin University of China, Beijing, China\\
$^2$Beijing Key Laboratory of Big Data Management and Analysis Methods, Beijing, China
\emails
\{lishaojie95, ouyangsheng, liuyonggsai\}@ruc.edu.cn
}

\begin{document}

\maketitle

\begin{abstract}
	The theoretical analysis of spectral clustering mainly focuses on consistency, while there is relatively little research on its generalization performance. In this paper, we study the excess risk bounds of the popular spectral clustering algorithms: \emph{relaxed} RatioCut and \emph{relaxed} NCut. Firstly, we show the convergence rate of their excess risk bounds between the empirical continuous optimal solution and the population-level continuous optimal solution. Secondly, we show the fundamental quantity in influencing the excess risk between the empirical discrete optimal solution and the population-level discrete optimal solution. At the empirical level, algorithms can be designed to reduce this quantity. Based on our theoretical analysis, we propose two novel algorithms that can not only penalize this quantity, but also cluster the out-of-sample data without re-eigendecomposition on the overall samples. Experiments verify the effectiveness of the proposed algorithms.
\end{abstract}


\section{Introduction}\label{section1}
\label{submission}
Spectral clustering is one of the most popular algorithms in unsupervised learning and has been widely used for many applications \cite{von2007tutorial,dhillon2001co,kannan2004clusterings,shaham2018spectralnet,liu2018global}. Given a set of data points independently sampled from an underlying unknown probability distribution, often referred to as the population distribution, spectral clustering algorithms aim to divide all data points into several disjoint sets based on some notion of similarity. Spectral clustering originates from the spectral graph partitioning \cite{fiedler1973algebraic}, and one way to understand spectral clustering is to view it as a relaxation of searching for the best graph-cut since the latter is known as an NP-hard problem \cite{von2007tutorial}. The core method of spectral clustering is the eigendecomposition on the graph Laplacian, and the matrix composed of eigenvectors can be interpreted as a lower-dimensional representation that preserves the grouping relationships among data points as much as possible. Subsequently, various methods such as $k$-means \cite{ng2001spectral,shi2000normalized}, dynamic programming \cite{alpert1995multiway}, or orthonormal transform \cite{stella2003multiclass} can be used to get the discrete solution on the matrix and therefore the final group partitions.

However, compared with the prosperous development of the design and application, the generalization performance analysis of spectral clustering algorithms appears to be not sufficiently well-documented. Hitherto, the theoretical analysis of spectral clustering mainly focuses on consistency \cite{von2008consistency,von2004convergence,cao2011consistency,trillos2018variational,trillos2016consistency,schiebinger2015geometry,terada2019kernel}. Consistency means that if it is true that as the sample size collected goes to infinity, the partitioning of the data constructed by spectral clustering converges to a certain meaningful partitioning on the population level \cite{von2008consistency}, but consistency alone does not indicate the sample complexity \cite{vapnik1999nature}. To our best knowledge, there is only one research that investigates the generalization performance of kernel NCut \cite{terada2019kernel}. They use the relationship between NCut and the weighted kernel $k$-means \cite{dhillon2007weighted}, based on which they establish the excess risk bounds for kernel NCut. However, their analysis focuses on the graph-cut solution, not the solution of spectral clustering that we used in practice. We leave more discussions about the related work in the Appendix.


Motivated by the above problems, we investigate the excess risk bound of the popular spectral clustering algorithms: \emph{relaxed} RatioCut and \emph{relaxed} NCut. To compare with the RatioCut and NCut that are without relaxation, we refer to spectral clustering as \emph{relaxed} RatioCut and \emph{relaxed} NCut in this paper. It is known that spectral clustering often consists of two steps \cite{von2007tutorial}: (1) to obtain the optimal continuous solution by the eigendecomposition on the graph Laplacian; (2) to obtain the optimal discrete solution, also referred to as discretization, from the continuous solution by some heuristic algorithms, such as $k$-means and orthonormal transform. Consistent with the two steps, we first investigate the excess risk bound between the empirical continuous optimal solution and the population-level continuous optimal solution. In deriving this bound, an immediate emerging difficulty is that the empirical continuous solution and the population-level continuous solution are in different dimensional spaces, making the empirical solution impossible to substitute into the expected error formula. To overcome this difficulty, we define integral operators, and use the spectral relationship between the integral operator and the graph Laplacian to extend the finite-dimensional eigenvector to the infinite-dimensional eigenfunction. Thus the deriving can proceed. We show that for both \emph{relaxed} RatioCut and \emph{relaxed} NCut, their excess risk bounds have a convergence rate of the order $\mathcal{O}(1/\sqrt{n})$. Secondly, we investigate the excess risk bound between the empirical discrete optimal solution and the population-level discrete optimal solution. We observe the fundamental quantity in influencing this excess risk, whose presence is caused by the heuristic algorithms used in step (2) of spectral clustering.  This fundamental quantity motivates us to design algorithms to penalize it from the empirical perspective, reducing it as small as possible. Meanwhile, we observe that the orthonormal transform \cite{stella2003multiclass} is an effective algorithm for penalizing this term, whose optimization objective corresponds to the empirical form of this fundamental quantity. Additionally, an obvious drawback of spectral clustering algorithms (\emph{relaxed} NCut and \emph{relaxed} RatioCut) is that they fail to generalize to the out-of-sample data points, requiring re-eigendecomposition on the overall data points. Based on our theoretical analysis, we propose two algorithms, corresponding to \emph{relaxed} NCut and \emph{relaxed} RatioCut, respectively, which can cluster the unseen samples without the eigendecomposition on the overall samples, largely reducing the time complexity. Moreover, when clustering the unseen samples, the proposed algorithms will penalize the fundamental quantity for searching for the optimal discrete solution, decreasing the excess risk. We have numerical experiments on the two algorithms, and the experimental results verify the effectiveness of our proposed algorithms. Our contributions are summarized as follows:
\begin{itemize}
	\item We provide the first excess risk bounds for the continuous solution of spectral clustering.
	\item We show the fundamental quantity in influencing the excess risk for the discrete solution of spectral clustering.  We then propose two algorithms that can not only penalize this term but also generalize to the new samples.
	\item The numerical experiments demonstrate the effectiveness of the proposed algorithms.
\end{itemize}


\section{Preliminaries}\label{section3}
In this section, we introduce some notations and have a brief introduction to spectral clustering. For more details, please refer to \cite{von2007tutorial}. 

Let $\mathcal{X}$ be a subset of $\mathbb{R}^d$, $\rho$ be a probability measure on $\mathcal{X}$, and $\rho_n$ be the empirical measure.
Given a set of samples $\mathbf{X} =  \{\mathbf{x}_1, \mathbf{x}_2, ... , \mathbf{x}_n\}$ independently drawn from the population distribution $\rho$, the weighted graph constructed on $\mathbf{X}$ can be specified by $\mathcal{G} = (\mathbb{V}, \mathbb{E}, \mathbf{W})$, where $\mathbb{V}$ denotes the set of all nodes, $\mathbb{E}$ denotes the set of all edges connecting the nodes, and $\mathbf{W}  := (\mathbf{W}_{i,j})_{n \times n} = (\frac{1}{n}W(\mathbf{x}_i,\mathbf{x}_j))_{n \times n}$ is a weight matrix calculated by the weight function $W(x,y)$. Let $|\mathbb{V}| = n $ denotes the number of all data points to be grouped. To cluster $n$ points into $K$ groups is to decompose $\mathbb{V}$ into $K$ disjoint sets, i.e., $\mathbb{V} = \cup^K_{l=1}\mathbb{V}_l$ and $\mathbb{V}_k \cap \mathbb{V}_l = \varnothing$,  $\forall k  \neq l$. We define the degree matrix $\mathbf{D}$ to be a diagonal matrix with entries $d_i = \sum_{j=1}^n \mathbf{W}_{i,j}$. Then, the unnormalized graph Laplacian is defined as $\mathbf{L} = \mathbf{D}-\mathbf{W}$, and the asymmetric normalized graph Laplacian is defined as $\mathbf{L}_{rw} = \mathbf{D}^{-1}\mathbf{L}=\mathbf{I}-\mathbf{D}^{-1}\mathbf{W}$.

We now present some facts about spectral clustering.
Let $\mathbf{U}=(\mathbf{u}_1,...,\mathbf{u}_K) \in \mathbb{R}^{n \times K}$, where $\mathbf{u}_1,...,\mathbf{u}_K$ are $K$ vectors. We define the following empirical error:
\begin{equation}\label{eq-3}
	\begin{aligned}
		\hat{F}(\mathbf{U}):=\frac{1}{2n(n-1)} \sum_{k=1}^K \sum_{i,j=1,i \neq j}^n \mathbf{W}_{i,j}(\mathbf{u}_{k,i}-\mathbf{u}_{k,j})^2,
	\end{aligned}
\end{equation}
where $\mathbf{u}_{k,i}$ means the $i$-th component of the $k$-th vector $\mathbf{u}_{k}$.
The optimization objective of RatioCut can be written as:
\begin{equation}\label{eq4}
	\begin{aligned}
		\min_{\mathbf{U}} \hat{F}(\mathbf{U})
		\text{ s.t. } \left \{\mathbf{u}_{i,j}= \frac{1}{\sqrt{|\mathbb{V}_j|}} \text{ if } v_i \in \mathbb{V}_j, \text{otherwise } 0 \right \},
	\end{aligned}
\end{equation}
where $|\mathbb{V}_j|$ denotes the number of vertices of a subset $\mathbb{V}_j$ of a graph.
The optimization objective of NCut can be written as:
\begin{equation}\label{eq5}
	\begin{aligned}
		\min_{\mathbf{U}}\hat{F}(\mathbf{U})
		\text{s.t.} \left \{\mathbf{u}_{i,j}=
		\frac{1}{\sqrt{\mathrm{vol}(\mathbb{V}_j)}} \text{if } v_i \in \mathbb{V}_j,\text{otherwise } 0\right\},
	\end{aligned}
\end{equation}
where $\mathrm{vol}(\mathbb{V}_j)$ denotes the summing weights of edges of a subset $\mathbb{V}_j$ of a graph. Since searching for the optimal solution of RatioCut and NCut is known as an NP-hard problem \cite{von2007tutorial}, spectral clustering often involves a relaxation operation, which allows the entries of $\mathbf{U}$ to take arbitrary real values \cite{von2007tutorial}. Thus the optimization objective of \emph{relaxed} RatioCut can be written as:
\begin{equation}\label{eq6}
	\begin{aligned}
		\min_{\mathbf{U}=(\mathbf{u}_1,...,\mathbf{u}_K)}\hat{F}(\mathbf{U}),\text{ s.t. }\mathbf{U}^T\mathbf{U} = \mathbf{I},
	\end{aligned}
\end{equation}
where $\mathbf{I}$ is the identity matrix.
The optimal solution of \emph{relaxed} RatioCut is given by choosing $\mathbf{U}$ as the matrix which contains the first $K$ eigenvectors of $\mathbf{L}$ as columns~\cite{von2007tutorial}.
Similarly, the optimization objective of \emph{relaxed} NCut can be written as:
\begin{equation}\label{eq7}
	\begin{aligned}
		\min_{\mathbf{U}=(\mathbf{u}_1,...,\mathbf{u}_K)}\hat{F}(\mathbf{U}),\text{ s.t. }
		\mathbf{U}^T\mathbf{D}\mathbf{U} = \mathbf{I}.
	\end{aligned}
\end{equation}
The optimal solution of \emph{relaxed} NCut is given by choosing the matrix $\mathbf{U}$ which contains the first $K$ eigenvectors of $\mathbf{L}_{rw}$ as columns \cite{von2007tutorial}.

\section{Excess Risk Bounds}\label{section4}
We consider the real space in this paper.
Let $W : \mathcal{X} \times \mathcal{X} \rightarrow R$ be a symmetric continuous weight function such that
\begin{equation}\label{eq8}
	0 < W(x,y) \leq C \quad x,y \in \mathcal{X},
\end{equation}
measuring the similarities between pairs of data points $x,y \in \mathcal{X}$. Since $W: \mathcal{X} \times \mathcal{X} \rightarrow R$ is not necessary to be positive definite and positive $W$ is more common in practice, we assume that $W$ to be positive in this paper. We now define the degree function as $m(x) = \int_\mathcal{X}W(x,y)d\rho(y)$, and then define the function: $p(x,y)=m(x)$ if $x = y$ and $0$ otherwise, which is the population counterpart of the degree matrix. Let $L^2(\mathcal{X},\rho)$ denotes the space of square integrable functions with norm $\| f \|^2_\rho = \langle f ,f \rangle_{\rho} = \int_{\mathcal{X}} |f(x)|^2 d\rho(x)$. 
\subsection{Relaxed RatioCut}\label{section41}
Based on the function $W$, we define the function $L: \mathcal{X} \times \mathcal{X} \rightarrow R$ 
\begin{equation} \nonumber
	L(x,y)= p(x,y) - W(x,y) \quad  x,y \in \mathcal{X},
\end{equation}
which is symmetric. When $L$ is restricted to $\forall \mathbf{X} =  \{\mathbf{x}_1, \mathbf{x}_2, ... , \mathbf{x}_n\}$ for any positive integer $n$, the corresponding matrix $\mathbf{L}$ is positive semi-definite (refer to proposition 1 in \cite{von2007tutorial}), thus $L(x,y)$ is a kernel function and associated with a reproducing kernel Hilbert space (RKHS) $\mathcal{H}$ with scalar product (norm) $\langle \cdot, \cdot \rangle $ ($\| \cdot \|$). In Section \ref{section41}, we assume $ \kappa = \sup_{x \in \mathcal{X}}L(x,x)$ and $L(x,y)$ to be continuous, which are common assumptions in spectral clustering. The elements in $\mathcal{H}$ are thus bounded continuous functions, and the corresponding integral operator $L_K:L^2(\mathcal{X},\rho) \rightarrow L^2(\mathcal{X},\rho)$
\begin{equation} \nonumber
	(L_Kf)(x) = \int_\mathcal{X} L(x,y)f(y)d\rho(y)
\end{equation}
is thus a bounded operator. The operator $L_K$ is the limit version of the Laplacian $\mathbf{L}$ \cite{rosasco2010learning}. In other words, the matrix $\mathbf{L}$ is an empirical version of the operator $L_K$.

To investigate the excess risk bound, we need to define the population-level error, a limit version of Eq. (\ref{eq-3}):
$$
F(U) := \frac{1}{2}\sum_{k=1}^K \iint W(x,y)(u_k(x)-u_k(y))^2d\rho(x)d\rho(y),
$$
where $U = (u_1,...,u_K)$ consists of $K$ functions $u_k$. Further, 
the optimization objective of the population-level error of \emph{relaxed} RatioCut, analogous to Eq. (\ref{eq6}), can be defined as:
\begin{align}\label{eq11}
	&\min_{U}F(U) 
	\text{  s.t.  } \langle u_i, u_j \rangle_{\rho} = 1 \text { if } i=j, \text{ otherwise }0.
\end{align}
Let $\widetilde{U}^* = (\widetilde{u}^*_1,...,\widetilde{u}^*_K)$ be the optimal solution of Eq. (\ref{eq11}). Actually, $\widetilde{u}^*_1,...,\widetilde{u}^*_K$ are eigenfunctions of the operator $L_K$ \cite{rosasco2010learning}, that is $L_K \widetilde{u}^*_k = \lambda_k(L_K) \widetilde{u}^*_k \; \mbox{for } k = 1,...,K$, where $\lambda_k(L_K)$ is an eigenvalue of the operator $L_K$, $k = 1,...,K$.

With the population-level error of \emph{relaxed} RatioCut, we begin to analyze the excess risk bound.
Excess risk measures on the population-level how the difference between the error of the empirical solution and the error of the population optima performs related to the sample size $n$ \cite{biau2008performance,liu2021refined,li2021sharper}, formalized as $F(\widetilde{\mathbf{U}}^*) - F(\widetilde{U}^*)$, where $\widetilde{\mathbf{U}}^* = (\widetilde{\mathbf{u}}^*_1,...,\widetilde{\mathbf{u}}^*_K)$ is the optimal solution of the empirical error of \emph{relaxed} RatioCut, i.e., Eq. (\ref{eq6}), and, actually, $\widetilde{\mathbf{u}}^*_1,...,\widetilde{\mathbf{u}}^*_K$ are the eigenvectors of Laplacian $\mathbf{L}$ \cite{von2007tutorial}.
However, an immediate difficulty to derive the bound of $F(\widetilde{\mathbf{U}}^*) - F(\widetilde{U}^*)$ is that $\widetilde{\mathbf{U}}^*$ and $\widetilde{U}^*$ are in different spaces. Specifically, $\widetilde{\mathbf{U}}^* \in \mathbb{R}^{n \times K}$ related to sample size $n$ is in finite-dimensional space, while $\widetilde{U}^*$ is in infinite-dimensional function space. The fact that for different sample size $n$, the elements in $\widetilde{\mathbf{U}}^*$ live in different spaces, making the term $F(\widetilde{\mathbf{U}}^*)$ impossible to be calculated.
To overcome this challenge,
we define operator $T_n: \mathcal{H}  \rightarrow \mathcal{H} $:
$$
T_n = \frac{1}{n} \sum_{i=1}^n \langle \cdot ,L_{\mathbf{x}_i} \rangle L_{\mathbf{x}_i},
$$
where $L_{\mathbf{x}_i} = L(\cdot,\mathbf{x}_i)$. And we denote $\check{U} = (\check{u}_1,...,\check{u}_K)$ as the first $K$ eigenfunctions of the operator $T_n$. \cite{rosasco2010learning} shows that $T_n$ and $\mathbf{L}$ have the same eigenvalues (up to zero eigenvalues) and their corresponding eigenfunctions and eigenvectors are closely related. If $\lambda_k$ is a nonzero eigenvalue and $\widetilde{\mathbf{u}}^*_k$, $\check{u}_k$ are the corresponding eigenvector and eigenfunction of $\mathbf{L}$ and $T_n$ (normalized to norm l in $\mathbb{R}^n$ and $\mathcal{H}$) respectively, then
\begin{equation}\label{eq13}
	\begin{aligned}
		\widetilde{\mathbf{u}}^*_k &= \frac{1}{\sqrt{\lambda_k}} \left(\check{u}_k(\mathbf{x}_1),...,\check{u}_k(\mathbf{x}_n) \right); \quad \\
		\check{u}_k(x) &= \frac{1}{\sqrt{\lambda_k}}\left(\frac{1}{n} \sum_{i=1}^n \widetilde{\mathbf{u}}^{*i}_kL(x, \mathbf{x}_i)\right),
	\end{aligned}
\end{equation}
where $\widetilde{\mathbf{u}}_k^{*i}$ is the $i$-th component of $\widetilde{\mathbf{u}}_k^*$.

From Eq. (\ref{eq13}), one can see that the eigenvectors of $\mathbf{L}$ are the empirical version of the eigenfunctions of $T_n$. In other words, if the eigenfunction $\check{u}_k(x)$ is restricted to the dataset $\mathbf{X}$, it can be mapped into the eigenvector $\widetilde{\mathbf{u}}^*_k$. Meanwhile, the eigenfunctions of $T_n$ are the extensions of the eigenvectors of $\mathbf{L}$, which are infinite-dimensional.
Back to the term $F(\widetilde{\mathbf{U}}^*) - F(\widetilde{U}^*)$, we can replace the vectors in $\widetilde{\mathbf{U}}^*$ by its corresponding extended eigenfunctions in $\check{U}$.
Therefore, we now can investigate the excess risk bound between the empirical continuous optimal solution and the population-level continuous optimal solution by bounding the term $ F(\check{U}) - F(\widetilde{U}^*)$.
Additionally, the relations between the eigenvectors in $\widetilde{\mathbf{U}}^*$ and the eigenfunctions in $\check{U}$ can be applied to cluster out-of-sample data points. One can approximately calculate the eigenvectors of the out-of-sample data by the eigenfunctions in $\check{U}$. Details will be shown in Section \ref{section5}. 
We now present the first excess risk bound for \emph{relaxed} RatioCut.
\begin{theorem}\label{theo1}
	Suppose for any $\check{u} \in \mathcal{H}$ such that $\|\check{u}\|_{\infty} \leq \sqrt{B}$, then for any $\delta > 0$, with probability at least $1-2\delta$, the term $ F(\check{U}) - F(\widetilde{U}^*)$ is upper bounded by
	$$
	8CBK\left(\sqrt{\frac{1}{n}}+2\sqrt{\frac{2\log \frac{1}{\delta}}{n}}\right)
	+K\frac{2 \kappa  \sqrt{2\log \frac{2}{\delta}} }{\sqrt{n}},
	$$
	where $C$ and $B$ are positive constants, $K$ is the clustering number.
\end{theorem}
\begin{remark}\rm{}
	Theorem \ref{theo1} shows that the excess risk bound of \emph{relaxed} RatioCut between the empirical continuous optimal solution and the population-level continuous optimal solution has a convergence rate of the order $\mathcal{O}\left(\frac{1}{\sqrt{n}} \right)$ if we assume that the eigenfunctions $\check{u} \in \mathcal{H}$ of operator $T_n$ are bounded, i.e., $\|\check{u}\|_{\infty} \leq \sqrt{B}$. This assumption is mild. 
	Since we assume the kernel function $L(x, y) \leq \kappa$ and is continuous, the elements
	in $\mathcal{H}$ associated with $L(x, y)$ are bounded. The definition
	of operator $T_n$ is: $ \mathcal{H} \rightarrow \mathcal{H}$, so it is reasonable to assume
	the eigenfunctions of $T_n$ are bounded, that is 
	$\|\check{u}\|_{\infty} \leq \sqrt{B}$. $C$ in Theorem \ref{theo1} comes from Eq. (\ref{eq8}).
	We provide the proof of Theorem \ref{theo1} in Appendix B.
\end{remark}
\begin{remark}\rm{}
	We highlight that we investigate the excess risk of spectral clustering. Compared with the generalization error bound $\hat{F}(\widetilde{\mathbf{U}}^*) - F(\widetilde{U}^*)$ that measures the difference between the empirical error of the empirical solution and the population-level error of the population-level solution, excess risk analysis is much more difficult because $\widetilde{\mathbf{U}}^*$ can not be calculated in expectation $F(\widetilde{\mathbf{U}}^*)$. The generalization error bound of \emph{relaxed} RatioCut is
	easier to obtain since $\widetilde{\mathbf{U}}^*$ can be directly substituted into $\hat{F}(\cdot)$ to calculate, and its proof  indeed is included in the proof of Theorem \ref{theo1}.  We show the generalization error bound as a corollary in the following.
\end{remark}
\begin{corollary}\label{coro1}
	Under the above assumptions, for any $\delta > 0$, with probability at least $1-\delta$,
		\begin{align*}
			\hat{F}(\widetilde{\mathbf{U}}^*) - F(\widetilde{U}^*) \leq
			K\frac{2 \sqrt{2} \kappa  \sqrt{\log \frac{2}{\delta}} }{\sqrt{n}},
		\end{align*}
		where $K$ is the clustering number.
\end{corollary}

In practice, after obtaining eigenvectors of the Laplacian $\mathbf{L}$, spectral clustering uses the heuristic algorithms on the eigenvectors to obtain the discrete solution. In analogy to this empirical process, we define the population-level discrete solution $\ddot{U} = (\ddot{u}_1,...,\ddot{u}_K)$, which are $K$ functions in RKHS $\mathcal{H}$ and are sought through $\mathcal{H}$ by the population-level continuous solution $\check{U}$. Let $U^* = (u^*_1,...,u^*_K)$ be the optimal solution of the minimal population-level error of RatioCut, i.e., optimal solution of the population-level version of Eq. (\ref{eq4}). We then investigate the excess risk between the empirical discrete optimal solution and the population-level discrete optimal solution by bounding the term $F(\ddot{U})-F(U^*)$.
\begin{theorem}\label{theo2}
	Suppose $\sum_{k=1}^K\|\ddot{u}_k-\check{u}_k\|_2 \leq \epsilon$ and for any $\check{u} \in \mathcal{H}$ such that $\|\check{u}\|_{\infty} \leq \sqrt{B}$, then for any $\delta > 0$, with probability at least $1-2\delta$, the term $F(\ddot{U})-F(U^*)$ is upper bounded by
	$$
	4C\epsilon + 8CBK\left(\sqrt{\frac{1}{n}}+2\sqrt{\frac{2\log \frac{1}{\delta}}{n}}\right)+\frac{2 K\kappa  \sqrt{2\log \frac{2}{\delta}} }{\sqrt{n}},
	$$  
	where $C$ and $B$ are positive constants, $K$ is the clustering number.
\end{theorem}
\begin{remark}\rm{}
	In the proof of Theorem \ref{theo2}, we make an error decomposition: $F(\ddot{U})-F(U^*) = \underset{\mathcal{A}}{\underbrace{F(\ddot{U}) - F(\check{U})}} + \underset{\mathcal{B}}{\underbrace{F(\check{U}) -\hat{F}(\widetilde{\mathbf{U}}^*)}} + \underset{\mathcal{C}}{\underbrace{\hat{F}(\widetilde{\mathbf{U}}^*) - F(\widetilde{U}^*)}}+ \underset{\mathcal{D}}{\underbrace{F(\widetilde{U}^*) - F(U^*)}}.$ Term $\mathcal{B}$ is proved by the empirical process theory, term $\mathcal{C}$ is proved by spectral properties of the integral operator and the operator theory, while term $\mathcal{D} \leq 0$ can be derived easily. Bounds of the terms $\mathcal{B}$ and $\mathcal{C}$ give the result of Theorem \ref{theo1}. For term $\mathcal{A}$, we show that it can be bounded by $4C\sum_{k=1}^K\|\ddot{u}_k-\check{u}_k\|_2$ (The proof is provided in Appendix C). We denote this quantity as $\epsilon$, and the upper bound reveals that $\sum_{k=1}^K\|\ddot{u}_k-\check{u}_k\|_2$ is a fundamental quantity in influencing the excess risk between the empirical discrete optimal solution and the population-level discrete optimal solution, which motivates us to penalize it as much as possible at the empirical level. We thus propose new algorithms in the next section. Additionally, since searching for the best graph-cut is known as an NP-hard problem \cite{von2007tutorial}, we investigate the generalization performance of the discrete solution obtained from the continuous solution in the practical spectral clustering process rather than the agnostic graph-cut solution. We hope that the theoretical study on such a kind of discrete solution can guide the design of novel spectral clustering algorithms.
\end{remark}

\subsection{Relaxed NCut}
The basic idea of this subsection is roughly the same as Section \ref{section41}. We consider \emph{relaxed} NCut corresponding to the asymmetric normalized Laplacian $\mathbf{L}_{rw}$. Bound (\ref{eq8}) implies the corresponding integral operator $\mathbb{L}:L^2(\mathcal{X},\rho) \rightarrow L^2(\mathcal{X},\rho)$
\begin{equation} \nonumber
	(\mathbb{L}f)(x) = f(x) - \int_\mathcal{X} \frac{W(x,y)f(y)}{m(x)}d\rho(y)
\end{equation}
is well defined and continuous. To avoid notations abuse, we use symbols provided in Section \ref{section41}. Corresponding minimal population-level error similar to Eq. (\ref{eq11}) can be easily written from the empirical version in Eq. (\ref{eq7}). For brevity, we omit it and just give some notations here. Let $\widetilde{U}^* = (\widetilde{u}^*_1,...,\widetilde{u}^*_K)$ be the optimal solution of the minimal population-level error of \emph{relaxed} NCut, which are eigenfunctions of the operator $\mathbb{L}$ \cite{rosasco2010learning}. We denote $\widetilde{\mathbf{U}}^* = (\widetilde{\mathbf{u}}^*_1,...,\widetilde{\mathbf{u}}^*_K)$ as the optimal solution of minimal empirical error of \emph{relaxed} NCut, i.e., Eq. (\ref{eq7}), which actually are eigenvectors of the Laplacian $\mathbf{L}_{rw}$\cite{von2007tutorial}.

Firstly, we aim to bound the term $F(\check{U}) - F(\widetilde{U}^*) $. However, another immediate difficulty is that the methods described in Section \ref{section41} are not directly applicable for \emph{relaxed} NCut. The operator corresponding to $T_n$ in the previous subsection appears to be impossible to be defined  for \emph{relaxed} NCut since $W$ is not necessarily positive definite, so there is no RKHS associated with it. Moreover, even if $W$ is positive definite, the operator $\mathbb{L}$ involves a division by a function, so there may not be a map from the RKHS $\mathcal{H}$ to itself. To overcome this challenge, we use an assumption on $W$ introduced in \cite{rosasco2010learning} to construct an auxiliary RKHS $\mathcal{H}$ associated with a continuous real-valued bounded kernel $\mathcal{K}$.
Here is the assumption:
\begin{assumption}\label{assu1}
	Assume that $W : \mathcal{X} \times \mathcal{X} \rightarrow R$ is a positive, symmetric function such that
	\begin{align*}
		W(x,y) \geq c > 0 \quad x,y \in \mathcal{X} ; 
		\quad W \in C_b^{d+1} (\mathcal{X} \times \mathcal{X}),
	\end{align*}
	where $C_b^{d+1}(\mathcal{X} \times \mathcal{X})$ is a family of continuous bounded functions such that all the (standard) deviations of orders exist and are continuous bounded functions. 
\end{assumption}
According to \cite{rosasco2010learning}, Assumption \ref{assu1} implies that there exists a RKHS $\mathcal{H}$ with bounded continuous kernel $\mathcal{K}$ such that: $W_{x},  \frac{1}{m_n}W_{x} \in \mathcal{H}$, where $W_x = W(\cdot, x)$ and $m_n = \frac{1}{n} \sum_{i = 1}^n W_{x_i}$.
This allows us to define the following empirical operators  $\mathbb{L}_n, A_n: \mathcal{H} \rightarrow \mathcal{H}$
	\begin{align*}
		A_n = \frac{1}{n} \sum_{i=1}^n \langle \cdot, \mathcal{K}_{\mathbf{x}_i} \rangle_{\mathcal{H}} \frac{1}{m_n} W_{\mathbf{x}_i}; 
		\quad \mathbb{L}_n = I-A_n,
	\end{align*}
where $\mathcal{K}_{x} = \mathcal{K}(\cdot, x)$. Let $\check{U} = (\check{u}_1,...,\check{u}_K)$ be the first $K$ eigenfunctions of the operator $\mathbb{L}_n$.
\cite{rosasco2010learning} shows that $\mathbb{L}_n$, $ A_n$ and $\mathbf{L}_{rw}$ have closely related eigenvalues and eigenfunctions. The spectra of $\mathbf{L}_{rw}$ and $\mathbb{L}_n$ are the same up to the eigenvalue $1$. Moreover,
if $\lambda_k  \neq 1$ is an eigenvalue and $\widetilde{\mathbf{u}}^*_k$, $\check{u}_k$ are the eigenvector and eigenfunction of $\mathbf{L}_{rw}$ and $\mathbb{L}_n$, respectively, then
\begin{equation}\label{eq15}
	\begin{aligned}
		\widetilde{\mathbf{u}}^*_k &= (\check{u}_k(\mathbf{x}_1),...,\check{u}_k(\mathbf{x}_n)); \\
		\check{u}_k(x) &= \frac{1}{1-\lambda_k} \frac{1}{n} \sum_{i=1}^n \frac{W(x,\mathbf{x}_i)}{m_n(x)} \widetilde{\mathbf{u}}^{*i}_k,
	\end{aligned}
\end{equation}
where $\widetilde{\mathbf{u}}^{*i}_k$ is the $i$-th component of the eigenvector $\widetilde{\mathbf{u}}^*_k $. From Eq. (\ref{eq15}), one can observe that the eigenvectors of $\mathbf{L}_{rw}$ are
the empirical version of the eigenfunctions of $\mathbb{L}_n$. Moreover, the eigenfunctions of $\mathbb{L}_n$ are the extensions of the eigenvectors of $\mathbf{L}_{rw}$, which are infinite-dimensional.
Therefore, given the eigenvectors of $\mathbf{L}_{rw}$, we can extend it to the corresponding eigenfunctions.
With this relationship, we can now investigate the excess risk between the empirical continuous optimal solution and the population-level continuous optimal solution by bounding the term $F(\check{U}) - F(\widetilde{U}^*)$. The following is the first theorem of \emph{relaxed} NCut.
\begin{theorem}\label{theo3}
	Under Assumption \ref{assu1}, suppose for any $\check{u} \in \mathcal{H}$ such that $\|\check{u}\|_{\infty} \leq \sqrt{B}$, then for any $\delta > 0$, with probability at least $1-2\delta$, the term $F(\check{U}) - F(\widetilde{U}^*)$ is upper bounded by 
	$$
	8CBK\left(\sqrt{\frac{1}{n}}+2\sqrt{\frac{2\log \frac{1}{\delta}}{n}}\right)  +KC\sqrt \frac{\log \frac{2}{\delta} }{n}.
	$$
	where $C$ and $B$ are positive constants, $K$ is the clustering number.
\end{theorem}
\begin{remark}\rm{}
	From Theorem \ref{theo3}, the excess risk of \emph{relaxed} NCut has a convergence rate of the order $\mathcal{O}\left(\frac{1}{\sqrt{n}}\right)$. The proof techniques used in Theorem \ref{theo3} conclude spectral properties of integral operators, operator theory, and empirical processes. $C$ in Theorem \ref{theo3} comes from Eq. (\ref{eq8}). 
	We provide the proof of Theorem \ref{theo3} in Appendix D. Moreover, the generalization error bound of \emph{relaxed} NCut is shown below.
	\begin{corollary}\label{coro2}
		Under the above assumptions, for any $\delta > 0$, with probability at least $1-\delta$,
			\begin{align*}
				\hat{F}(\widetilde{\mathbf{U}}^*) - F(\widetilde{U}^*) \leq
				KC\sqrt \frac{\log \frac{2}{\delta} }{n},
			\end{align*}
			where $K$ is the clustering number.
	\end{corollary}
\end{remark}

As discussed before, the continuous solution of spectral clustering typically involves a discretization process, thus we then investigate the excess risk bound between the empirical discrete optimal solution
and the population-level discrete optimal solution for \emph{relaxed} NCut. In analogy to the previous subsection, we investigate $F(\ddot{U})-F(U^*)$, where $\ddot{U} = (\ddot{u}_1,...,\ddot{u}_K$) are $K$ functions in RKHS $\mathcal{H}$ and are sought through $\mathcal{H}$ by the continuous eigenfunctions $\check{U}$, and where $U^* = (u^*_1,...,u^*_K)$ is the optimal solution of
the minimal population-level error of NCut, i.e., optimal solution of the population-level version of Eq. (\ref{eq5}).
\begin{theorem}\label{theo4}
	Under Assumption \ref{assu1}, suppose $\sum_{k=1}^K\|\ddot{u}_k-\check{u}_k\|_2 \leq \epsilon$ and for any $\check{u} \in \mathcal{H}$ such that $\|\check{u}\|_{\infty} \leq \sqrt{B}$, then for any $\delta > 0$, with probability at least $1-2\delta$, the term $F(\ddot{U})-F(U^*)$ is upper bounded by 
	$$
	4C\epsilon + 8CBK\left(\sqrt{\frac{1}{n}}+2\sqrt{\frac{2\log \frac{1}{\delta}}{n}}\right)+KC\sqrt \frac{\log \frac{2}{\delta} }{n}
	$$
	where $C$ and $B$ are positive constants, $K$ is the clustering number.
\end{theorem}
\begin{remark}\rm{}
	From Theorem 4, one can see that the term $\sum_{k=1}^K\|\ddot{u}_k-\check{u}_k\|_2$ is also a fundamental quantity in influencing the excess risk of \emph{relaxed} NCut between the empirical discrete optimal solution and the population-level discrete optimal solution, which motivates us to propose algorithms in the next section to penalize this term to make the risk bound as small as possible.
	In addition to the difficulties mentioned above,  proving excess risk bounds also has the following difficulties: (1) the objective function of spectral clustering (see Eq (\ref{eq-3})) is a pairwise function, which can not be written as a summation of independent and identically distributed (i.i.d.) random variables so that the standard techniques in the i.i.d. case can not apply to it. In this paper, we use the $U$-process technique introduced in \cite{clemenccon2008ranking} to overcome this difficulty. 
	(2) the operator $\mathbb{L}$ involves a division by a function, thus the term $\mathcal{C}$ can not be bounded directly by the proof technique of Theorem \ref{theo2}. We must introduce equivalent probability measures to construct equivalent vector space (please refer to Appendix D). 
\end{remark}
\begin{remark}\rm{}
	This remark discusses why we use the asymmetric normalized Laplacian, not the symmetric normalized Laplacian. Using the asymmetric normalized graph Laplacian, we can analyze \emph{relaxed} NCut in a unified form of the empirical error (i.e., Eq. (\ref{eq-3})). While for the normalized symmetric Laplacian, we need to transform Eq. (\ref{eq-3}) to 
	\begin{align*}
		\hat{F}(\mathbf{U}):=\frac{1}{2n(n-1)} \sum_{k=1}^K \sum_{i,j=1,i \neq j}^n \mathbf{W}_{i,j}\left(\frac{\mathbf{u}_{k,i}}{\sqrt{d_i}}-\frac{\mathbf{u}_{k,j}}{\sqrt{d_j}}\right)^2.
	\end{align*} Please refer to Proposition 3 and Eq. (11) in \cite{von2007tutorial} for details. 
\end{remark}
\begin{remark}\rm{}
	This remark discusses the relationship between this paper and \cite{li2021sharper}. \cite{li2021sharper} study the clustering algorithm through a general framework and then gives excess risk bounds based on this framework. Specifically, the excess risk in \cite{li2021sharper} is of the form $F(\widetilde{\mathbf{U}}^*) - F(\widetilde{U}^*)$. However, we have discussed that $F(\widetilde{\mathbf{U}}^*)$ is impossible to be calculated for spectral clustering due to the dimensional issue. Thus, the bounds of $F(\widetilde{\mathbf{U}}^*) - F(\widetilde{U}^*)$ established in \cite{li2021sharper} do not hold for the specific spectral clustering problem, and that's also the reason why we introduce the integral operator tool to revisit the spectral clustering problem. Hence, we highlight that the results of this paper, both the bounds and the algorithms, are novel compared to \cite{li2021sharper}.
\end{remark}
\section {Algorithms}\label{section5}
From Theorems \ref{theo2} and \ref{theo4}, the imperative is to penalize $\sum_{k=1}^K\|\ddot{u}_k-\check{u}_k\|_2$ to make it as small as possible. Towards this aim, we should solve the following formula to find the optimal discrete solution $\ddot{U}$:
$$
\ddot{U} := \underset{U = (u_1,...,u_K)}{\arg \min} \sum_{k=1}^K\|u_k-\check{u}_k\|_2 \quad
\mbox{s.t.} \; u_k(x) \in \{0,1\},
$$
where $U = (u_1,...,u_K)$ is any set of $K$ functions in RKHS $\mathcal{H}$. In the corresponding empirical clustering process, we should optimize this term $\sum_{k=1}^K\|\ddot{\mathbf{u}}_k-\widetilde{\mathbf{u}}^*_k\|_2$. It can be roughly equivalent to optimize $\|\ddot{\mathbf{U}}-\widetilde{\mathbf{U}}^*\|_F$, to find the optimal discrete solution $\ddot{\mathbf{U}} = (\ddot{\mathbf{u}}_1,...,\ddot{\mathbf{u}}_K)$, where $F$ denotes the Frobenius norm. \cite{stella2003multiclass} propose an iterative fashion to optimize $\|\ddot{\mathbf{U}}-\widetilde{\mathbf{U}}^*\|_F$ to get the discrete solution closest to the continuous optimal solution $\widetilde{\mathbf{U}}^*$. At a high level, this paper provides a theoretical explanation on \cite{stella2003multiclass} from the population view.

The idea in \cite{stella2003multiclass} is based on that the continuous optimal solutions consist of not only the eigenvectors but of a whole family spanned by the eigenvectors through orthonormal transform. Thus the discrete optimal solution can be searched by orthonormal transform. With this idea, we can solve the following optimization objective to find the optimal discrete solution $\ddot{\mathbf{U}}$ and orthonormal transform:
\begin{equation}\nonumber
	\begin{aligned}
		&(\ddot{\mathbf{U}},\mathbf{R}^*) := \underset{\mathbf{U},\mathbf{R}}{\arg \min}{\|\mathbf{U}-\widetilde{\mathbf{U}}^*\mathbf{R}\|} \quad \\
		\mbox{s.t.}\quad
		&\mathbf{U} \in \{0,1\}^{n\times K},
		\mathbf{U}\mathbf{1}_K=\mathbf{1}_n,
		\mathbf{R}\mathbf{R}^T=\mathbf{I}_K,
	\end{aligned}
\end{equation}
where $\mathbf{1}_n$ is a vector with all one elements, $\mathbf{U}$ is any set of $K$ discrete vectors in the eigenspace, and $\mathbf{R} \in \mathbb{R}^{K \times K}$ is an orthonormal matrix.
The orthonormal transform program finds the optimal discrete solution in an iterative fashion. This  iterative fashion is shown below:

\noindent(1) given $\mathbf{R}^*$,
solving the following formula:
$$
\underset{\mathbf{U}}{\arg \min}{\|\mathbf{U}-\widetilde{\mathbf{U}}^*\mathbf{R}^*\|}, \quad
\mbox{s.t. } \;
\mathbf{U} \in \{0,1\}^{n\times K},
\mathbf{U}\mathbf{1}_K=\mathbf{1}_n.
$$

\noindent(2) given $\ddot{\mathbf{U}}$, solving the following formula:
$$
\underset{\mathbf{R}}{\arg \min}{\|\ddot{\mathbf{U}}-\widetilde{\mathbf{U}}^*\mathbf{R}\|}, \quad
\mbox{s.t.} \;
\mathbf{R}\mathbf{R}^T=\mathbf{I}_K.
$$
We denote this iterative fashion in \cite{stella2003multiclass} as $\mathrm{POD}$ (Program of Optimal Discretization).
\subsection{GPOD}
We now introduce our proposed algorithms, called $Generalized$ $\mathrm{POD}$ $(\mathrm{GPOD})$ algorithm, which can not only penalize the fundamental quantity in influencing the excess risk of the discrete solution but also allow clustering the unseen data points. 

Firstly,
for the samples $\mathbf{X}$, we can use the eigenvectors $\widetilde{\mathbf{U}}^*$ of $\mathbf{L}$ (or $\mathbf{L}_{rw}$) to obtain its extensions based on Eq. (\ref{eq13}) (or Eq. (\ref{eq15})), that is to obtain the eigenfunctions $\check{U}$ of $T_n$ (or $\mathbb{L}_n$). 
Secondly, when the new data points $\bar{\mathbf{X}} = \{\bar{\mathbf{x}}_1,...,\bar{\mathbf{x}}_m\}$ come, we can calculate its eigenvectors $\bar{\mathbf{U}} = \{\bar{\mathbf{u}}_1,...,\bar{\mathbf{u}}_K\}\in \mathbb{R}^{m \times K}$ with the help of the eigenfunctions $\check{U}= (\check{u}_1,...,\check{u}_K)$. By mapping the eigenfunctions into finite dimensional space, we can approximately obtain the eigenvectors of the new samples $\bar{\mathbf{X}}$. Specifically, we can use formula 
\begin{align*}
	\bar{\mathbf{u}}_k = \frac{1}{\sqrt{\lambda_k}} (\check{u}_k(\bar{\mathbf{x}}_1),...,\check{u}_k(\bar{\mathbf{x}}_m))
\end{align*} 
to obtain the eigenvectors of $\bar{\mathbf{X}}$ for \emph{relaxed} RatioCut and use 
\begin{align*}
	\bar{\mathbf{u}}_k = (\check{u}_k(\bar{\mathbf{x}}_1),...,\check{u}_k(\bar{\mathbf{x}}_m)
\end{align*}
for \emph{relaxed} NCut.
Note that for \emph{relaxed} RatioCut, since the underlying $\rho$ is unknown, the term $L(x, \mathbf{x}_i)$ can be empirically approximated by $\frac{1}{n} \sum_{i =1}^n W(\cdot, \mathbf{x}_i) - W(\cdot, \mathbf{x}_i)$.

After obtaining the eigenvectors of the out-of-sample data points $\bar{\mathbf{X}}$, we can use the $\mathrm{POD}$ iterative fashion to optimize the following optimization problem to seek the empirical optimal discrete solution:
	\begin{align*}
		&(\ddot{\mathbf{U}},\mathbf{R}^*) := \underset{\mathbf{U},\mathbf{R}}{\arg \min}{\|\mathbf{U}-\bar{\mathbf{U}}\mathbf{R}\|} \quad\\
		\mbox{s.t.} \quad
		&\mathbf{U} \in \{0,1\}^{m\times K},
		\mathbf{U}\mathbf{1}_K=\mathbf{1}_m,
		\mathbf{R}\mathbf{R}^T=\mathbf{I}_K.
	\end{align*}
	This optimization process can penalize the fundamental quantity for the out-of-sample data points.
	
The ability of our proposed algorithm in clustering unseen data points  without the eigendecomposition on the overall data points makes the spectral clustering more applicable, largely reducing the time complexity. The concrete algorithm steps are presented in Appendix F, where we also analyze how the time complexity of our proposed algorithm is significantly improved in Remark 1. Overall, the proposed algorithms can not only penalize the fundamental quantity but also cluster the out-of-sample data points.
\begin{remark}\rm{}
	Eqs. (\ref{eq13}) and (\ref{eq15}) hold when the denominator is not $0$. This remark discusses the case when the denominator is $0$, i.e., the $0$ or $1$ eigenvalue. According to the spectral projection view, for the unnormalized Laplacian, respectively the asymmetric graph Laplacian, the $0$-eigenvalue, respectively the $1$ eigenvalue, doesn't affect the performance of spectral clustering, see Proposition 9 and Proposition 14 in \cite{rosasco2010learning}, respectively.
	Thus, the $0$ or $1$ eigenvalue doesn't influence the performance of $\mathrm{GPOD}$ in clustering the out-of-sample data.
\end{remark}
	\subsection{Numerical Experiments}
	We have numerical experiments on the two proposed algorithms. Considering the length limit, we leave the experimental settings and results in Appendix G. The experimental results show that the proposed algorithms can cluster the out-of-sample data points, verifying their effectiveness. 
	
	\section {Conclusions}
	In this paper, we investigate the generalization performance of popular spectral clustering algorithms: \emph{relaxed} RatioCut and \emph{relaxed} Ncut, and provide the excess risk bounds. According to the two steps of practical spectral clustering algorithms, we first provide a convergence rate of the order $\mathcal{O}\left(1/\sqrt{n}\right)$ for the continuous solution for both \emph{relaxed} RatioCut and \emph{relaxed} Ncut. We then show the fundamental quantity in influencing the excess risk of the discrete solution. Theoretical analysis inspires us to propose two novel algorithms that can not only cluster the out-of-sample data, largely reducing the time complexity, but also penalize this fundamental quantity to be as small as possible. By numerical experiments, we verify the effectiveness of the proposed algorithms. One limitation of this paper is that we don't provide a true  convergence rate for the excess risk of the empirical discrete solution. We believe that this problem is pretty important and worthy of further study. 
	
	\bibliographystyle{named}
	\bibliography{ijcai22}
	
	\onecolumn
	\maketitle
	\appendix
	\section {Related Work}
	This section introduces related work on the theoretical analysis of spectral clustering algorithms.
	Existing theoretical research of spectral clustering mainly focuses on consistency, i.e., if it is true that as the sample size collected goes to infinity, the partitioning of the data constructed by the clustering algorithm converges to a certain meaningful partitioning on the population level. \cite{von2008consistency} establishes consistency for the embedding by proving that as much as the eigenvectors of the Laplacian matrix converge uniformly to the eigenfunctions of the Laplacian operator. \cite{rosasco2010learning} provides the simpler proof of this convergence. \cite{cao2011consistency} constructs the consistency of regularized spectral clustering. \cite{rohe2011spectral} analyzes the consistency for stochastic block models, \cite{ting2011analysis} analyzes the spectral convergence, \cite{pelletier2011operator} analyzes the convergence of graph Laplacian, and \cite{singer2017spectral} analyzes the convergence of the connection graph Laplacian. \cite{trillos2016consistency} proposes a framework and improves the results in \cite{arias2012normalized} by minimizing the discrete functionals over all possible partitions of the data points, while the latter just minimizes a specific family of subsets of the data points. Based on the framework in \cite{trillos2016consistency}, \cite{trillos2018variational} provides a variational approach known as $\Gamma$-convergence, proving the convergence of the spectrum of the graph Laplacian towards the spectrum of a corresponding continuous operator. \cite{terada2019kernel} investigates the kernel normalized cut, establishing the consistency by the weighted $k$-means on the reproducing kernel Hilbert space (RKHS), and deriving the excess risk bound for kernel NCut. However, as we discussed in the main paper, they study the graph-cut solution, not the solution of spectral clustering that we used in practice.	Different from the above research, we investigate the excess risk bound of the popular spectral clustering algorithms (\emph{relaxed} RatioCut and \emph{relaxed} NCut), not consistency. Our analysis is based on the practical steps of spectral clustering and spans two perspectives: the continuous solution and the discrete solution.

	\section{Proof of Theorem 1}
	\begin{proof}
		The term $F(\check{U})-F(\widetilde{U}^*)$ can be decomposed as:
		\begin{equation} \nonumber
			\begin{aligned}
				F(\check{U})-F(\widetilde{U}^*)& = \underset{\mathcal{B}}{\underbrace{F(\check{U}) -\hat{F}(\widetilde{\mathbf{U}}^*)}} + \underset{\mathcal{C}}{\underbrace{\hat{F}(\widetilde{\mathbf{U}}^*) - F(\widetilde{U}^*)}}.
			\end{aligned}
		\end{equation}

		\noindent$\textbf{(1).}$ For term $\mathcal{B}$, we have
		\begin{equation} \nonumber
			\begin{aligned}
				\mathcal{B} = F(\check{U}) - \hat{F}(\widetilde{\mathbf{U}}^*) 
				= \frac{1}{2}\sum_{k=1}^K \left(\iint W(x,y)(\check{u}_k(x)-\check{u}_k(y))^2d\rho(x)d\rho(y) -\frac{1}{n(n-1)}\sum_{i,j=1,i \neq j}^n \mathbf{W}_{i,j}(\widetilde{\mathbf{u}}^*_{k,i}-\widetilde{\mathbf{u}}^*_{k,j})^2 \right).
			\end{aligned}
		\end{equation}
		Based on Eq. (8) in the main paper and by the transformation of elements between the RKHS and $L^2(\mathcal{X},\rho)$, we have
		\begin{equation} \nonumber
			\begin{aligned}
				\mathcal{B}  = \frac{1}{2}\sum_{k=1}^K  \left(\iint W(x,y)(\check{u}_k(x)-\check{u}_k(y))^2d\rho(x)d\rho(y)  -\frac{1}{n(n-1)}\sum_{i,j=1,i \neq j}^n \mathbf{W}_{i,j}(\check{u}_k(\mathbf{x}_i)-\check{u}_k(\mathbf{x}_j))^2  \right).
			\end{aligned}
		\end{equation}
		The term $\mathcal{B}$ can be equivalently written as
		\begin{equation} \nonumber
			\begin{aligned}
				\mathcal{B} = &\frac{1}{2}\sum_{k=1}^K \left (\mathbb{E} \left[W(x,y)(\check{u}_k(x)-\check{u}_k(y))^2\right]-\hat{\mathbb{E}} \left [W(x,y)(\check{u}_k(x)-\check{u}_k(y))^2 \right] \right ),
			\end{aligned}
		\end{equation}
		where $\mathbb{E}$ denotes the expectation and $\hat{\mathbb{E}}$ denotes the corresponding empirical average. Furthermore, denoted by $\ell_{\check{u}_k}(x,y) =W(x,y)(\check{u}_k(x)-\check{u}_k(y))^2$.
		For any $\check{u}$ in RKHS $\mathcal {H}$,
		the term $\mathcal{B}$ can be upper bounded by $$\frac{1}{2}K\sup_{\check{u} \in \mathcal{H}} (\mathbb{E}[\ell_{\check{u}}] -  \hat{\mathbb{E}}[\ell_{\check{u}}]).$$
		
		We first apply the McDiarmid's inequality \cite{mohri2018foundations} to control the deviation of the term $\sup_{\check{u} \in \mathcal{H}} (\mathbb{E}[\ell_{\check{u}}] -  \hat{\mathbb{E}}[\ell_{\check{u}}])$ from its expectation.
		For independent and identically distributed (i.i.d.) sampled data points $\mathbf{X} = \{\mathbf{x}_1,...,\mathbf{x}_{t-1},\mathbf{x}_t,\mathbf{x}_{t+1},...,\mathbf{x}_n\}$ and $\bar{\mathbf{X}} = \{\mathbf{x}_1,...,\mathbf{x}_{t-1},\bar{\mathbf{x}}_t,\mathbf{x}_{t+1},...,\mathbf{x}_n\}$, we have
		\begin{equation} \nonumber
			\begin{aligned}
				& \hphantom{{}={}}\left | \underset{\check{u}\in \mathcal{H}} {\sup}  (\mathbb{E}[\ell_{\check{u}}] -  \hat{\mathbb{E}}_\mathbf{X}[\ell_{\check{u}}])- \underset{\check{u}\in \mathcal{H}} {\sup}  (\mathbb{E}[\ell_{\check{u}}]-  \hat{\mathbb{E}}_{\bar{\mathbf{X}}}[\ell_{\check{u}}])  \right|\\
				&\leq \underset{\check{u}\in \mathcal{H}} {\sup}  \left|\hat{\mathbb{E}}_\mathbf{X}[\ell_{\check{u}}]-  \hat{\mathbb{E}}_{\bar{\mathbf{X}}}[\ell_{\check{u}}] \right | \\
				&\leq \frac{2}{n(n-1)}\underset{\check{u}\in \mathcal{H}} {\sup} \sum_{j=1,j\neq t}^n (|\ell_{\check{u}}(\mathbf{x}_t,\mathbf{x}_j)|+|\ell_{\check{u}}(\bar{\mathbf{x}}_t,\mathbf{x}_j)|)\\
				&\leq \frac{4}{n}\underset{\check{u}\in \mathcal{H}} {\sup} \| \ell_{\check{u}}\|_\infty.\\
			\end{aligned}
		\end{equation}
		Since
		we assume $\|\check{u}\|_{\infty} \leq \sqrt{B}$, thus $\underset{x,y}{\sup} (\check{u}(x)-\check{u}(y))^2 \leq 4B$. Together with $W(x,y) \leq C$ gives
		\begin{equation} \nonumber
			\begin{aligned}
				&\left| \underset{\check{u}\in \mathcal{H}} {\sup}  (\mathbb{E}[\ell_{\check{u}}] -  \hat{\mathbb{E}}_\mathbf{X}[\ell_{\check{u}}])- \underset{\check{u}\in \mathcal{H}} {\sup}  (\mathbb{E}[\ell_{\check{u}}]-  \hat{\mathbb{E}}_{\bar{\mathbf{X}}}[\ell_{\check{u}}]) \right| \leq \frac{16}{n}CB.\\
			\end{aligned}
		\end{equation}
		Applying McDiarmid's inequality with increment bounded by $\frac{16}{n}CB$ implies that with probability at least $1-\delta$, we have
		\begin{equation}  \nonumber
			\begin{aligned}
				\underset{\check{u}\in \mathcal{H}} {\sup}  (\mathbb{E}[\ell_{\check{u}}]-\hat{\mathbb{E}}_\mathbf{X}[\ell_{\check{u}}])\leq 	\mathbb{E}\underset{\check{u}\in \mathcal{H}} {\sup}  (\mathbb{E}[\ell_{\check{u}}] -  \hat{\mathbb{E}}_\mathbf{X}[\ell_{\check{u}}])+ 32CB\sqrt{\frac{2\log \frac{1}{\delta}}{n}}.
			\end{aligned}
		\end{equation}
		We use the Rademacher average \cite{bartlett2002rademacher} to bound the term $\mathbb{E}\underset{\check{u}\in \mathcal{H}} {\sup}  (\mathbb{E}[\ell_{\check{u}}]-\hat{\mathbb{E}}_\mathbf{X}[\ell_{\check{u}}])$. As mentioned in the main paper, the objective function of spectral clustering, i.e., Eq. (1), is a pairwise function, which can not be written as a summation of i.i.d. random variables, so that the standard techniques in the i.i.d. case can not apply to it. We use the $U$-process technique introduced in \cite{clemenccon2008ranking} to overcome this difficulty. Specifically, we define the following Rademacher complexity for spectral clustering:
		\begin{definition}
			Assume $\mathcal{H}$ is a space of functions $\check{u}$, then the empirical Rademacher complexity of $\mathcal{H}$ for spectral clustering is:
				\begin{align*}
					\hat{R}_n (\mathcal{H})= \mathbb{E}_{\sigma} \left[\sup_{\check{u} \in \mathcal{H}} \left| \frac{2}{\lfloor n/2 \rfloor} \sum_{i = 1}^{\lfloor n/2 \rfloor} \sigma_i W(\mathbf{x}_i,\mathbf{x}_{i + \lfloor n/2 \rfloor})
					(\check{u}(\mathbf{x}_i) - \check{u}(\mathbf{x}_{i + \lfloor n/2 \rfloor}))^2 \right| \right],
				\end{align*}
			where $\sigma_1,...,\sigma_{\lfloor n/2 \rfloor}$ is an i.i.d. family of Rademacher variables taking values $-1$ and $1$ with equal probability independent of the sample $\mathbf{X}$, and $\lfloor n/2 \rfloor$  is the largest integer no greater than $\frac{n}{2}$. The Rademacher complexity of $\mathcal{H}$ is $R(\mathcal{H}) = \mathbb{E} \hat{R}_n (\mathcal{H})$.
		\end{definition}
		With the Rademacher complexity, we begin to bound the term $\mathbb{E}\underset{\check{u}\in \mathcal{H}} {\sup}  (\mathbb{E}[\ell_{\check{u}}]-\hat{\mathbb{E}}_\mathbf{X}[\ell_{\check{u}}])$. Lemma A.1 in \cite{clemenccon2008ranking} with $\mathbf{q}_{\check{u}} (\mathbf{x}_i,\mathbf{x}_j) = \mathbb{E} [\ell_{\check{u}}]- \ell_{\check{u}}(\mathbf{x}_i,\mathbf{x}_j)$ and the index set $\mathcal{H}$ allow us to derive $$\mathbb{E}\underset{\check{u}\in \mathcal{H}} {\sup}  \left(\mathbb{E}[\ell_{\check{u}}]-\hat{\mathbb{E}}_\mathbf{X}[\ell_{\check{u}}] \right) \leq \mathbb{E}\underset{\check{u}\in \mathcal{H}} {\sup} \left[\mathbb{E}[\ell_{\check{u}}] - \frac{1}{\lfloor \frac{n}{2} \rfloor} \sum_{i=1}^{\lfloor \frac{n}{2} \rfloor} \ell_{\check{u}} (\mathbf{x}_i,\mathbf{x}_{\lfloor \frac{n}{2} \rfloor+i}) \right].$$
		Let $\mathbf{X}' = \{\mathbf{x}'_1,...,\mathbf{x}'_n\}$ be i.i.d. samples independent of $\mathbf{X}$ and let $\{\sigma\}_{i \in [\lfloor \frac{n}{2} \rfloor]} $ be a sequence of Rademacher variables. According to the Jensen's inequality and a standard symmetrization technique, the term $\mathbb{E}_{\mathbf{X}}\underset{\check{u}\in \mathcal{H}} {\sup}  \left(\mathbb{E}[\ell_{\check{u}}]-\hat{\mathbb{E}}_\mathbf{X}[\ell_{\check{u}}] \right)$ can be bounded by
		\begin{equation} \nonumber
			\begin{aligned}
				&\mathbb{E}_{\mathbf{X},\mathbf{X}'} \underset{\check{u}\in \mathcal{H}} {\sup}  \frac{1}{\lfloor \frac{n}{2} \rfloor}\left [\sum_{i=1}^{\lfloor \frac{n}{2} \rfloor} \ell_{\check{u}} (\mathbf{x}'_i,\mathbf{x}'_{\lfloor \frac{n}{2} \rfloor+i}) - \sum_{i=1}^{\lfloor \frac{n}{2} \rfloor} \ell_{\check{u}} (\mathbf{x}_i,\mathbf{x}_{\lfloor \frac{n}{2} \rfloor+i})\right ]\\
				=& \mathbb{E}_{\mathbf{X},\mathbf{X}',\sigma} \underset{\check{u}\in \mathcal{H}} {\sup}  \frac{1}{\lfloor \frac{n}{2} \rfloor} \left [\sum_{i=1}^{\lfloor \frac{n}{2} \rfloor} \sigma_i \left( \ell_{\check{u}} (\mathbf{x}'_i,\mathbf{x}'_{\lfloor \frac{n}{2} \rfloor+i}) -  \ell_{\check{u}} (\mathbf{x}_i,\mathbf{x}_{\lfloor \frac{n}{2} \rfloor+i}) \right) \right ]\\
				=& \frac{2}{\lfloor \frac{n}{2}\rfloor}  \mathbb{E}_{\mathbf{X},\sigma} \underset{\check{u}\in \mathcal{H}} {\sup} \sum_{i=1}^{\lfloor \frac{n}{2} \rfloor} \sigma_i  \ell_{\check{u}} (\mathbf{x}_i,\mathbf{x}_{\lfloor \frac{n}{2} \rfloor+i})\\
				\leq& \frac{2}{\lfloor \frac{n}{2}\rfloor}  \mathbb{E}_{\mathbf{X},\sigma}  \left|\underset{\check{u}\in \mathcal{H}} {\sup} \sum_{i=1}^{\lfloor \frac{n}{2} \rfloor} \sigma_i  \ell_{\check{u}} (\mathbf{x}_i,\mathbf{x}_{\lfloor \frac{n}{2} \rfloor+i}) \right |\\ \quad
				\leq& \frac{2}{\lfloor \frac{n}{2}\rfloor}  \mathbb{E}_{\mathbf{X}}  \left(\underset{\check{u}\in \mathcal{H}} {\sup} \sum_{i=1}^{\lfloor \frac{n}{2} \rfloor} [ \ell_{\check{u}} (\mathbf{x}_i,\mathbf{x}_{\lfloor \frac{n}{2} \rfloor+i})]^2 \right )^{1/2},
			\end{aligned}
		\end{equation}
		where the last inequality uses the Khinchin-Kahane inequality \cite{latala1994best}.
		Since $\underset{x,y}{\sup} (\check{u}(x)-\check{u}(y))^2 \leq 4B$ and $W(x,y) \leq C$, thus we can bound the last formula by $\frac{8BC}{\lfloor \frac{n}{2}\rfloor} \sqrt{{\lfloor \frac{n}{2} \rfloor}}  
		\leq 16BC\sqrt{\frac{1}{n} }$.
		Based on the above results, we derive that the term $\mathcal{B}$ can be bounded by $ 8CBK \left(\sqrt{\frac{1}{n}}+2\sqrt{\frac{2\log \frac{1}{\delta}}{n}} \right)$ with probability at least $1-\delta$.
		
		\noindent$\textbf{(2).}$
		To bound the term $\mathcal{C}$, we need to define another operator: $T_\mathcal{H}: \mathcal{H}  \rightarrow \mathcal{H} $ for \emph{relaxed} RatioCut:
		\begin{align*}
			&T_\mathcal{H} = \int_\mathcal{X} \langle \cdot ,L_x\rangle L_x d\rho(x),
		\end{align*}
		where $L_x = L(\cdot,x)$. \cite{rosasco2010learning} shows that $T_\mathcal{H}$ and $L_K$ have the same eigenvalues (possibly up to some zero eigenvalues) and their corresponding eigenfunctions are closely related. A similar relation holds for $T_n$ and $\mathbf{L}$ that we have mentioned in the main paper. The spectral properties between the operators and the Laplacian can help us to bound the term $\mathcal{C}$.
		
		According to the spectral properties of the graph Laplacian, we know that $\hat{F}(\widetilde{\mathbf{U}}^*)$  is equivalent to the first $K$ smallest eigenvalues of $\mathbf{L}$ \cite{von2007tutorial}. Similarly, with operator spectral properties, $F(\widetilde{U}^*)$ is equivalent to the first $K$ smallest eigenvalues of operator $L_K$. Specifically, $F(\widetilde{U}^*)$ can be written as:
		\begin{align*}
			F(\widetilde{U}^*) &= \frac{1}{2}\sum_{k=1}^K \iint W(x,y)(\widetilde{u}^*_k(x)-\widetilde{u}^*_k(y))^2d\rho(x)d\rho(y)\\
			&= \sum_{k=1}^K \int  \widetilde{u}^*_k(x) L_K \widetilde{u}^*_k(x) d\rho(x) = \sum_{k=1}^K \langle \widetilde{u}^*_k, L_K\widetilde{u}^*_k \rangle_{\rho}= \sum_{k=1}^K \lambda_k(L_K) \langle \widetilde{u}^*_k, \widetilde{u}^*_k \rangle_{\rho} = \sum_{k=1}^K \lambda_k(L_K),
		\end{align*}
		where $\lambda_k(L_K)$ is the $k$-th eigenvalue of the operator $L_K$. Thus, for term $\mathcal{C}$, we have:
		\begin{equation} \nonumber
			\begin{aligned}
				\mathcal{C} &= \hat{F}(\widetilde{\mathbf{U}}^*) - F(\widetilde{U}^*) = \sum_{i=1}^K  \lambda_i(\mathbf{L})- \lambda_i(L_{K}). \\
			\end{aligned}
		\end{equation}
		According to Proposition 8 and Proposition 9 in \cite{rosasco2010learning} that demonstrates the relationship of eigenvalues between operator $L_K$ and operator $T_{\mathcal{H}}$, operator $T_n$ and matrix $\mathbf{L}$, respectively, we thus obtain that $\mathcal{C} = \sum_{i=1}^K  \lambda_i(T_n)- \lambda_i(T_{\mathcal{H}}) $. Furthermore, it can be bounded by $K \sup_j|{\lambda}_j(T_n)- \lambda_j(T_{\mathcal{H}})|$. Since $T_n$ and $T_{\mathcal{H}}$ are self-joint operators \cite{rosasco2010learning}, from Theorem 5 in \cite{kato1987variation}, we can bound $\sup_j |\lambda_j(T_n)- \lambda_j(T_{\mathcal{H}})|$ by $\|T_n- T_{\mathcal{H}}\|$. Using the operator Theory \cite{lang2012real}, $\|T_n- T_{\mathcal{H}}\| \leq \|T_n- T_{\mathcal{H}}\|_{HS}$. From Theorem 7 in \cite{rosasco2010learning}, we know that $\|T_n- T_{\mathcal{H}}\|_{HS} \leq \frac{2 \sqrt{2} \kappa  \sqrt{\log \frac{2}{\delta}} }{\sqrt{n}}$ with probability at least $1-\delta$.
		From the above results, the term $\mathcal{C}$ can be bounded by $K\frac{2 \sqrt{2} \kappa  \sqrt{\log \frac{2}{\delta}} }{\sqrt{n}}$ with probability at least $1-\delta$.
		Based on the bounds of term $\mathcal{B}$ and term $\mathcal{C}$, we derive that $F(\check{U}) - \hat{F}(\widetilde{\mathbf{U}}^*)  \leq  8CBK(\sqrt{\frac{1}{n}}+2\sqrt{\frac{2\log \frac{1}{\delta}}{n}}) +K\frac{2 \sqrt{2} \kappa  \sqrt{\log \frac{2}{\delta}} }{\sqrt{n}}$ with probability at least $1-2\delta$.
	\end{proof}

	\section{Proof of Theorem 2}
	\begin{proof}
		The term $F(\ddot{U})-F(U^*)$ can be decomposed as:
		\begin{equation} \nonumber
			\begin{aligned}
				F(\ddot{U})-F(U^*)& = \underset{\mathcal{A}}{\underbrace{F(\ddot{U}) - F(\check{U})}} + \underset{\mathcal{B}}{\underbrace{F(\check{U}) -\hat{F}(\widetilde{\mathbf{U}}^*)}} + \underset{\mathcal{C}}{\underbrace{\hat{F}(\widetilde{\mathbf{U}}^*) - F(\widetilde{U}^*)}}+ \underset{\mathcal{D}}{\underbrace{F(\widetilde{U}^*) - F(U^*)}}.
			\end{aligned}
		\end{equation}

		\noindent$\textbf{(1).}$ Suppose $\sum_{k=1}^K\|\ddot{u}_k-\check{u}_k\|_2 \leq \epsilon$,
		\begin{equation}\nonumber
			\begin{aligned}
				&\mathcal{A}  = F(\ddot{U}) - F(\check{U}) \\
				= &\frac{1}{2}\sum_{k=1}^K \iint W(x,y)(\ddot{u}_k(x)-\ddot{u}_k(y))^2d\rho(x)d\rho(y) - \frac{1}{2}\sum_{k=1}^K \iint W(x,y)(\check{u}_k(x)-\check{u}_k(y))^2d\rho(x)d\rho(y) \\
				= &\frac{1}{2}\sum_{k=1}^K \iint \left (W(x,y)((\ddot{u}_k(x)-\ddot{u}_k(y))^2 -(\check{u}_k(x)-\check{u}_k(y))^2) \right)d\rho(x)d\rho(y) \\
				= &\frac{1}{2}\sum_{k=1}^K \iint W(x,y) \left((\ddot{u}_k(x)-\ddot{u}_k(y)+\check{u}_k(x)-\check{u}_k(y)) (\ddot{u}_k(x)-\ddot{u}_k(y)-\check{u}_k(x)+\check{u}_k(y)) \right)d\rho(x)d\rho(y)   \\
				\leq &\frac{1}{2}C\sum_{k=1}^K \iint \Big(\underset{q(x)}{\underbrace{(\ddot{u}_k(x)+\check{u}_k(x))}}-\underset{q(y)}{\underbrace{(\ddot{u}_k(y)+\check{u}_k(y))}}\Big) \times \Big(\underset{p(x)}{\underbrace{(\ddot{u}_k(x)-\check{u}_k(x))}}-\underset{p(y)}{\underbrace{(\ddot{u}_k(y)-\check{u}_k(y))}}\Big)d\rho(x)d\rho(y)\\
				= &\frac{1}{2}C\sum_{k=1}^K \iint \left(p(x)q(x)-p(y)q(x) -p(x)q(y)+p(y)q(y)\right) d\rho(x)d\rho(y)\\
				= &\frac{1}{2}C\sum_{k=1}^K \left(\int p(x)q(x)d\rho(x)-\iint p(y)q(x)d\rho(x)d\rho(y) -\iint p(x)q(y)d\rho(x)d\rho(y)+\int p(y)q(y)d\rho(y) \right) \\
				= &C\sum_{k=1}^K \left(\int p(x)q(x)dx-\int p(x)d\rho(x)\int q(x)d\rho(x) \right)\\
				\leq &C\sum_{k=1}^K \left(\int |p(x)q(x)|dx+\int |p(x)|d\rho(x)\int |q(x)|d\rho(x) \right)\\
				\leq &C\sum_{k=1}^K [\underset{\mbox{(use H\"{o}lder inequality)}}{\underbrace{ \left(\int |p(x)|^2dx \right)^{1/2}  \left(\int |q(x)|^2dx \right)^{1/2}}}+\int |p(x)|d\rho(x) \left(\int |\ddot{u}_k(x)|d\rho(x) + \int |\check{u}_k(x)|d\rho(x) \right) ]\\
				\leq &C\sum_{k=1}^K  [ \left(\int |p(x)|^2dx \right)^{1/2}  \underset{\mbox{(use Minkowski inequality)}}{\underbrace{[ \left(\int |\ddot{u}_k(x)|^2dx \right)^{1/2} + \left(\int |\check{u}_k(x)|^2dx \right)^{1/2} ] }} +\int |p(x)|d\rho(x) \left (\int |\ddot{u}_k(x)|d\rho(x)+\int |\check{u}_k(x)|d\rho(x) \right) ] \\
				= &C\sum_{k=1}^K \left(\|p\|_2  (\|\ddot{u}_k\|_2+\|\check{u}_k\|_2)+\|p\|_1(\|\ddot{u}_k\|_1 + \|\check{u}_k\|_1)\right)\\
				\leq  &C\sum_{k=1}^K \Big(\|p\|_2  (\|\ddot{u}_k\|_2+\|\check{u}_k\|_2)+\underset{\mbox{(use Lyapunov inequality)}}{\underbrace{\|p\|_2(\|\ddot{u}_k\|_2 + \|\check{u}_k\|_2)}} \Big).\\
			\end{aligned}
		\end{equation}
				
		Since the eigenfunction $\check{u}_k$ is normalized to norm $1$, and the discrete solution $\ddot{u}_k$ is also constrained to norm $1$, so the term $\mathcal{A}$ can be bounded by
		$4C\sum_{k=1}^K \|p\|_2 $. And since $\sum_{k=1}^K\|\ddot{u}_k-\check{u}_k\|_2 \leq \epsilon$, we finally bound $\mathcal{A}$ by $4C\epsilon$.
		
		\noindent$\textbf{(2).}$ Term $\mathcal{B}$ and Term $\mathcal{C}$ have been bounded in the proof of Theorem 1.
		
		\noindent$\textbf{(3).}$ $\widetilde{U}^*$ is the continuous solution and $U^*$ is the discrete solution. The continuous solution space is a larger solution space, so we obtain $\mathcal{D} =  F(\widetilde{U}^*) - F(U^*) \leq 0$.
		
		Based on the above results of $\mathcal{A}$, $\mathcal{B}$, $\mathcal{C}$, and $\mathcal{D}$, we derive that $F(\ddot{U})-F(U^*) \leq 4C\epsilon + 8CBK(\sqrt{\frac{1}{n}}+2\sqrt{\frac{2\log \frac{1}{\delta}}{n}}) +K\frac{2 \sqrt{2} \kappa  \sqrt{\log \frac{2}{\delta}} }{\sqrt{n}}$ with probability at least $1-2\delta$.
	\end{proof}
	
	\section{Proof of Theorem 3}
	\begin{proof}
		
		
		
		
		
		The term $F(\check{U})-F(\widetilde{U}^*)$ can be decomposed as:
		\begin{equation} \nonumber
			\begin{aligned}
				F(\check{U})-F(\widetilde{U}^*)& = \underset{\mathcal{B}}{\underbrace{F(\check{U}) -\hat{F}(\widetilde{\mathbf{U}}^*)}} + \underset{\mathcal{C}}{\underbrace{\hat{F}(\widetilde{\mathbf{U}}^*) - F(\widetilde{U}^*)}}.
			\end{aligned}
		\end{equation}
		
		\noindent$\textbf{(1).}$
		For term $\mathcal{B}$, we have
		\begin{equation} \nonumber
			\begin{aligned}
				\mathcal{B} = F(\check{U}) - \hat{F}(\widetilde{\mathbf{U}}^*) 
				= \frac{1}{2}\sum_{k=1}^K \left(\iint W(x,y)(\check{u}_k(x)-\check{u}_k(y))^2d\rho(x)d\rho(y) -\frac{1}{n(n-1)}\sum_{i,j=1,i \neq j}^n \mathbf{W}_{i,j}(\widetilde{\mathbf{u}}^*_{k,i}-\widetilde{\mathbf{u}}^*_{k,j})^2\right).
			\end{aligned}
		\end{equation}
		Based on Eq. (9) in the main paper, we have
		\begin{equation} \nonumber
			\begin{aligned}
				\mathcal{B}  = \frac{1}{2}\sum_{k=1}^K \left(\iint W(x,y)(\check{u}_k(x)-\check{u}_k(y))^2d\rho(x)d\rho(y)  -\frac{1}{n(n-1)}\sum_{i,j=1,i \neq j}^n \mathbf{W}_{i,j}(\check{u}_k(\mathbf{x}_i)-\check{u}_k(\mathbf{x}_j))^2 \right).
			\end{aligned}
		\end{equation}
		The term $\mathcal{B}$ can be equivalently written as
		\begin{equation} \nonumber
			\begin{aligned}
				\mathcal{B} = &\frac{1}{2}\sum_{k=1}^K \left (\mathbb{E}[W(x,y)(\check{u}_k(x)-\check{u}_k(y))^2] -\hat{\mathbb{E}}[W(x,y)(\check{u}_k(x)-\check{u}_k(y))^2] \right ),
			\end{aligned}
		\end{equation}
		where $\mathbb{E}$ is the expectation and $\hat{\mathbb{E}}$ is the corresponding empirical average.
		Denoted by $\ell_{\check{u}_k} =W(x,y)(\check{u}_k(x)-\check{u}_k(y))^2$.
		For any $\check{u}$ in the RKHS $\mathcal {H}$,
		the term $\mathcal{B}$ can be bounded by $$\frac{1}{2}K\sup_{\check{u} \in \mathcal{H}} (\mathbb{E}[\ell_{\check{u}}] -  \hat{\mathbb{E}}[\ell_{\check{u}}]).$$
		Till here, the following proof is the same as the proof of Theorem 1. For brevity, we omit it here.
		Since we assume $\|\check{u}\|_{\infty} \leq \sqrt{B}$, we have $\underset{x,y}{\sup} (\check{u}(x)-\check{u}(y))^2 \leq 4B$. Together with $W(x,y) \leq C$ gives that with probability at least $1-\delta$
		\begin{equation} \nonumber
			\begin{aligned}
				\mathcal{B} \leq 8CBK \left(\sqrt{\frac{1}{n}}+2\sqrt{\frac{2\log \frac{1}{\delta}}{n}} \right).
			\end{aligned}
		\end{equation}
		
		\noindent$\textbf{(2).}$
		To bound the term $\mathcal{C}$, we also need to define the following bounded operators $\mathbb{L_{\mathcal{H}}}, A_{\mathcal{H}} : \mathcal{H} \rightarrow \mathcal{H}$ for \emph{relaxed} NCut:
		\begin{equation} \nonumber
			\begin{aligned}
				&A_{\mathcal{H}} = \int_\mathcal{X} \langle \cdot, \mathcal{K}_x \rangle_{\mathcal{H}} \frac{1}{m} W_x d\rho(x), \\
				&\mathbb{L}_{\mathcal{H}} = I-A_{\mathcal{H}}.
			\end{aligned}
		\end{equation}
		\cite{rosasco2010learning} shows that $\mathbb{L}_{\mathcal{H}}$,  $A_{\mathcal{H}} $ and $\mathbb{L}$ have closely related eigenvalues and eigenfunctions, and the similar relations hold for $\mathbb{L}_n$, $ A_n$ and $\mathbf{L}_{rw}$. The spectral properties of these integral operators can help us to derive the term $\mathcal{C}$.
		
		Similar to the proof of Theorem 1, the next key step is to prove the value of $\hat{F}(\widetilde{\mathbf{U}}^*) $ and $ F(\widetilde{U}^*)$.
		We first introduce a measure $\rho_W = m\rho$, having density $m$ w.r.t $\rho$, is equivalent to $\rho$ since they have the same null sets. This implies that the spaces $L^2(\mathcal{X},\rho)$ and $L^2(\mathcal{X},\rho_W)$ are the same vector space and the corresponding norm are equivalent. In this proof, we regard $\mathbb{L}$ as an operator from $L^2(\mathcal{X},\rho_W)$ to $L^2(\mathcal{X},\rho_W)$, observing that its eigenvalues and eigenfunctions are the same as eigenvalues and eigenfunctions of $\mathbb{L}$, viewed as an operator from $L^2(\mathcal{X},\rho)$ into $L^2(\mathcal{X},\rho)$ \cite{rosasco2010learning}. Let $f \in L^2(\mathcal{X}, \rho_W)$,
		\begin{equation} \nonumber
			\begin{aligned}
				\langle \mathbb{L}f, f  \rangle_{\rho_W} &= \int_\mathcal{X} |f(x)|^2m(x)d\rho (x)- \int_\mathcal{X} \left(\int_\mathcal{X} \frac{W(x,s)}{m(x)}f(s)d\rho (s) \right) f(x)m(x)d\rho(x)\\
				&= \frac{1}{2}\int_\mathcal{X}\int_\mathcal{X} \left[|f(x)|^2W(x,s) - 2W(x,s)f(x)f(s) + |f(x)|^2W(x,s) \right]d\rho(s)d\rho(x)\\
				&= \frac{1}{2}\int_\mathcal{X}\int_\mathcal{X} W(x,s)|f(x)-f(s)|^2d\rho(s)d\rho(x)\\
				&= \langle L_K f, f  \rangle_{\rho},
			\end{aligned}
		\end{equation}
		where the last equality is obtained because eigenvalues and eigenfunctions of $\mathbb{L}$ are the same in $L^2(\mathcal{X},\rho)$ and $L^2(\mathcal{X}, \rho_w)$ \cite{rosasco2010learning}. So for the \emph{relaxed} NCut, $F(\widetilde{U}^*) = \sum_{k=1}^K \langle  L_K\widetilde{u}^*_k , \widetilde{u}^*_k\rangle_{\rho} = \sum_{k=1}^K \langle \mathbb{L} \widetilde{u}^*_k, \widetilde{u}^*_k \rangle_{\rho_W}$, which is equal to the sum of the first $K$ smallest eigenvalues of $\mathbb{L}$. Similarly, by replacing $\rho$ with the empirical measure $\frac{1}{n}\sum_{i=1}^n \delta_{x_i}$, $\hat{F}(\widetilde{\mathbf{U}}^*)$ is equal to the the sum of the first $K$ smallest eigenvalues of $\mathbf{L}_{rw}$ (This result can also be obtained by the spectral properties of $\mathbf{L}_{rw}$ \cite{von2007tutorial}). Then we have
		\begin{equation} \nonumber
			\begin{aligned}
				\mathcal{C} &= \hat{F}(\widetilde{\mathbf{U}}^*) - F(\widetilde{U}^*)\\
				& = \sum_{i=1}^K  \lambda_i(\mathbf{L}_{rw}) - \lambda_i(\mathbb{L})   \\
				&= \sum_{i=1}^K   \lambda_i(\mathbb{L}_n) - \lambda_i(\mathbb{L}_{\mathcal{H}})  \quad \mbox{(use Proposition 13 and Proposition 14 in \cite{rosasco2010learning})} \\
				&\leq K \underset{j} {\sup} |\lambda_j(\mathbb{L}_n)- \lambda_j(\mathbb{L}_{\mathcal{H}})| \\
				&\leq K \|\mathbb{L}_n- \mathbb{L}_{\mathcal{H}}\| \quad \mbox{(use Theorem 1 in \cite{kato1987variation})} \\
				&\leq K\|\mathbb{L}_n- \mathbb{L}_{\mathcal{H}}\|_{HS} \quad \mbox{(use operator theory in \cite{lang2012real})} \\
				&= K\|A_n- A_{\mathcal{H}} \|_{HS}\\
				& \leq KC \frac{\sqrt{\log \frac{2}{\delta}}}{\sqrt{n}} \quad \mbox{(use Theorem 15 in \cite{rosasco2010learning})}\\
			\end{aligned}
		\end{equation}
		with probability at least $1-\delta$.
		
		Based on the above results, we derive that $F(\check{U})-F(\widetilde{U}^*) \leq 8CBK(\sqrt{\frac{1}{n}}+2\sqrt{\frac{2\log \frac{1}{\delta}}{n}}) +KC\sqrt \frac{\log \frac{2}{\delta} }{n}$ with probability at least $1-2\delta$.
		
	\end{proof}
	
	
	
	
	
	
	\section{Proof of Theorem 4}
	\begin{proof}
		Similarly, the term $F(\ddot{U})-F(U^*)$ can be decomposed as:
		\begin{equation} \nonumber
			\begin{aligned}
				F(\ddot{U})-F(U^*)& = \underset{\mathcal{A}}{\underbrace{F(\ddot{U}) - F(\check{U})}} + \underset{\mathcal{B}}{\underbrace{F(\check{U}) -\hat{F}(\widetilde{\mathbf{U}}^*)}} + \underset{\mathcal{C}}{\underbrace{\hat{F}(\widetilde{\mathbf{U}}^*) - F(\widetilde{U}^*)}}+ \underset{\mathcal{D}}{\underbrace{F(\widetilde{U}^*) - F(U^*)}}.
			\end{aligned}
		\end{equation}
		
		\noindent$\textbf{(1).}$
		The proof of the term $\mathcal{A}$ is the same as the proof of Theorem 1.
		Suppose $\sum_{k=1}^K\|\ddot{u}_k-\check{u}_k\|_2 \leq \epsilon$,
		then we have
		\begin{equation} \nonumber
			\begin{aligned}
				\mathcal{A}  = &F(\ddot{U}) - F(\check{U}) \leq 4C \epsilon.
			\end{aligned}
		\end{equation}
		
		\noindent$\textbf{(2).}$ The term $\mathcal{B}$ and term $\mathcal{C}$ have been bounded in the proof of Theorem 3.
		
		\noindent$\textbf{(3).}$ It is easily to have $\mathcal{D} = F(\widetilde{U}^*) - F(U^*) \leq 0$.
		
		Based on the above results, we derive that $F(\ddot{U})-F(U^*) \leq 4C\epsilon + 8CBK(\sqrt{\frac{1}{n}}+2\sqrt{\frac{2\log \frac{1}{\delta}}{n}}) +KC\sqrt \frac{\log \frac{2}{\delta} }{n}$ with probability at least $1-2\delta$.
	\end{proof}
	\section{Algorithms}
	Algorithm \ref{alg:example1} corresponds to \emph{relaxed} RatioCut, while Algorithm \ref{alg:example2} corresponds to \emph{relaxed} NCut. We just show the pseudocode of clustering the out-of-sample data points in Algorithms \ref{alg:example1} and \ref{alg:example2}.  For clustering the original data $\mathbf{X}$, one can use the algorithm $\mathrm{POD}$. We provide the pseudocode of $\mathrm{POD}$ in Algorithm \ref{alg:example3} \cite{stella2003multiclass}, please refer to \cite{stella2003multiclass} for more details. Additionally, line $8$ in Algorithms \ref{alg:example1} and \ref{alg:example2} aims to normalize the length of each row of the matrix $\bar{\mathbf{U}}$ so that they lie on a unit hypersphere centered at the origin and then can be searched for the discrete solution through orthonormal transform when performing $\mathrm{POD}(\hat{\mathbf{U}})$ in line 9. Moreover, in line $8$, $\mathrm{Diag}$ denotes vector diagonalization operation and $\mathrm{diag}$ returns the diagonal of its matrix argument in a column vector. The following iterative fashion $\mathrm{POD}(\hat{\mathbf{U}})$ in line 9 aims to find the empirical optimal discrete solution $\ddot{\mathbf{U}}$ and the right orthonormal transform $\mathbf{R}^*$, see the details in Algorithm \ref{alg:example3}. 
	
	\begin{algorithm}[H]
		\caption{$\mathrm{GPOD}$ (\emph{relaxed} RatioCut)}
		\label{alg:example1}
		\textbf{Input:} weight function $W$, cluster number $K$, samples $\mathbf{X} =  \{\mathbf{x}_1, \mathbf{x}_2, ... , \mathbf{x}_n\}$, new samples $\bar{\mathbf{X}}= \{\bar{\mathbf{x}}_1, \bar{\mathbf{x}}_2, ... , \bar{\mathbf{x}}_m\}$.\\
		\textbf{Phase 1:} Based on $\mathbf{X}$, compute $\mathbf{L}$=$\mathbf{D}$-$\mathbf{W}$, then
		compute the smallest $K$ eigenvalues $\lambda_{i=1}^K$ and the corresponding eigenvectors $\widetilde{\mathbf{U}}^* = (\widetilde{\mathbf{u}}_1^{*},...,\widetilde{\mathbf{u}}_K^{*})$.\\
		\textbf{Phase 2:} Compute the eigenvectors of $\bar{\mathbf{X}}$, then find the optimal discrete solution $\ddot{\mathbf{U}}$ by the following steps:\\
		\vspace{-0.4cm}
		\begin{algorithmic}[1]
			\FOR{$k=1$ {\bfseries to} $K$}
			\FOR{$i=1$ {\bfseries to} $m$}
			\STATE  
			calculate $s_n(\bar{\mathbf{x}}_i)= \frac{1}{n}\sum_{j =1}^nW(\bar{\mathbf{x}}_i, \mathbf{x}_j)$ \\
			calculate $\hat{s}_n(\bar{\mathbf{x}}_i) =  \frac{1}{n}\sum_{j =1}^n (s_n(\bar{\mathbf{x}}_i) - W(\bar{\mathbf{x}}_i, \mathbf{x}_j)) \widetilde{\mathbf{u}}_k^{*j} $\\
			calculate $\bar{\mathbf{u}}_k(\bar{\mathbf{x}}_i) = \frac{1}{\sqrt{\lambda_k}}\hat{s}_n(\bar{\mathbf{x}}_i)$\\
			\ENDFOR
			\STATE concatenate $\bar{\mathbf{u}}_k = (\bar{\mathbf{u}}_k(\bar{\mathbf{x}}_1),...,\bar{\mathbf{u}}_k(\bar{\mathbf{x}}_m))$
			\ENDFOR
			\STATE concatenate  $\bar{\mathbf{U}} = (\bar{\mathbf{u}}_1,...,\bar{\mathbf{u}}_K)$
			\STATE normalize $\hat{\mathbf{U}}= \mathrm{Diag}(\mathrm{diag}^{-\frac{1}{2}}(\bar{\mathbf{U}} \bar{\mathbf{U}}^{T}))\bar{\mathbf{U}}$
			\STATE perform $\mathrm{POD}(\hat{\mathbf{U}})$, output $(\ddot{\mathbf{U}},\mathbf{R}^*)$.
		\end{algorithmic}
	\end{algorithm}
	
	\begin{algorithm}[H]
		\caption{$\mathrm{GPOD}$ (\emph{relaxed} NCut)}
		\label{alg:example2}
		\textbf{Input:} weight function $W$, cluster number $K$, samples $\mathbf{X} =  \{\mathbf{x}_1, \mathbf{x}_2, ... , \mathbf{x}_n\}$, new samples $\bar{\mathbf{X}}= \{\bar{\mathbf{x}}_1, \bar{\mathbf{x}}_2, ... , \bar{\mathbf{x}}_m\}$.\\
		\textbf{Phase 1:} Based on $\mathbf{X}$, compute $\mathbf{L}_{rw} = \mathbf{I} - \mathbf{D}^{-1} \mathbf{W}$, then
		compute the smallest $K$ eigenvalues $\lambda_{i=1}^K$ and the corresponding eigenvectors $\widetilde{\mathbf{U}}^* = (\widetilde{\mathbf{u}}_1^{*},...,\widetilde{\mathbf{u}}_K^{*})$.\\
		\textbf{Phase 2:} Compute the eigenvectors of $\bar{\mathbf{X}}$, then find the optimal discrete solution $\ddot{\mathbf{U}}$ by the following steps:\\
		\vspace{-0.4cm}
		\begin{algorithmic}[1]
			\FOR{$k=1$ {\bfseries to} $K$}
			\FOR{$i=1$ {\bfseries to} $m$}
			\STATE 
			calculate $s_n(\bar{\mathbf{x}}_i)= \frac{1}{n}\sum_{j =1}^nW(\bar{\mathbf{x}}_i, \mathbf{x}_j)$. \\
			calculate $\hat{s}_n(\bar{\mathbf{x}}_i) =  \frac{1}{n}\sum_{j =1}^n ( \frac{ W(\bar{\mathbf{x}}_i, \mathbf{x}_j}{s_n(\bar{\mathbf{x}}_i)}) \widetilde{\mathbf{u}}_k^{*j}$\\
			calculate $\bar{\mathbf{u}}_k(\bar{\mathbf{x}}_i) = \frac{1}{1 - \lambda_k}\hat{s}_n(\bar{\mathbf{x}}_i)$  \\
			\ENDFOR
			\STATE concatenate $\bar{\mathbf{u}}_k = (\bar{\mathbf{u}}_k(\bar{\mathbf{x}}_1),...,\bar{\mathbf{u}}_k(\bar{\mathbf{x}}_m))$
			\ENDFOR
			\STATE concatenate  $\bar{\mathbf{U}} = (\bar{\mathbf{u}}_1,...,\bar{\mathbf{u}}_K)$
			\STATE normalize $\hat{\mathbf{U}}= \mathrm{Diag}(\mathrm{diag}^{-\frac{1}{2}}(\bar{\mathbf{U}} \bar{\mathbf{U}}^{T}))\bar{\mathbf{U}}$
			\STATE perform $\mathrm{POD}(\hat{\mathbf{U}})$, output $(\ddot{\mathbf{U}},\mathbf{R}^*)$.
		\end{algorithmic}
	\end{algorithm}
	
	\begin{remark}\rm{} \textbf{[Comparison of time complexity.]}
		For the original $\mathrm{POD}$, the time complexity is mainly spent on the eigendecomposition and the SVD, whose complexity is all of the order $\mathcal{O}(n^3)$ \cite{stella2003multiclass}. If the iterative fashion in $\mathrm{POD}$ is performed $t$ times for $n$ samples, the time complexity of $\mathrm{POD}$ is of the order $\mathcal{O}((1+t)n^3)$, because $\mathrm{POD}$ needs to compute $1$ time eigendecomposition and $t$ times SVD. Therefore, when the out-of-sample data points $\bar{\mathbf{X}}$ come, assuming the iteration fashion in $\mathrm{POD}$ is $t$ times on the overall data points of size $n+m$, the time complexity of $\mathrm{POD}$ is $\mathcal{O}((1+t)(n+m)^3)$, because they need to compute $1$ time eigendecomposition and $t$ times SVD on the overall data points. While for our proposed algorithm, assuming the iteration fashion in $\mathrm{POD}$ is $t_1$ times for samples $\bar{\mathbf{X}}$, the time complexity is $\mathcal{O}((1+t_1)m^3)$, because we can calculate the eigenvectors of $\bar{\mathbf{X}}$ with the help of the extended eigenfunctions $\check{U}= (\check{u}_1,...,\check{u}_K)$, as discussed in the main paper. We just need to compute $1$ time eigendecomposition and $t_1$ times SVD on the out-of-sample data points $\bar{\mathbf{X}}$. Besides, the size $m$ of $\bar{\mathbf{X}}$ may be not large in practice, thus the term $\mathcal{O}((1+t_1)m^3)$ will be much smaller than $\mathcal{O}((1+t)(n+m)^3)$. Furthermore, a smaller $m$ may lead to faster convergence, thus $t_1$ may be much smaller than $t$ in practice. Based on the above analysis, one can see that our proposed algorithms significantly improve the time complexity when clustering unseen samples.
	\end{remark}
	\begin{algorithm}[H]
		\caption{$\mathrm{POD}$}
		\label{alg:example3}
		\textbf{Input:} matrix $\hat{\mathbf{U}}$.
		\begin{algorithmic}[1]
			\STATE  Initialize $\ddot{\mathbf{U}}$ by computing $\mathbf{R}^*$ as:\\
			$\mathbf{R}_1^* = [\hat{\mathbf{U}}(i,1),...,\hat{\mathbf{U}}(i,K)]^T$, random $i \in [n]$\\
			$c=0_{n \times 1}$\\
			For $k=2,...,K$, do:\\
			\quad $c=c+abs(\hat{\mathbf{U}} \mathbf{R}_{k-1}^*)$\\
			\quad $\mathbf{R}_k^* = [\hat{\mathbf{U}}(i,1),...,\hat{\mathbf{U}}(i,K)]^T$, $i = \arg \min c$
			\STATE  initialize convergence monitoring parameter $\bar{\phi}^* = 0$.
			\STATE find the optimal discrete solution $\ddot{\mathbf{U}}$ by:\\
			\quad  $\tilde{\mathbf{U}}= \hat{\mathbf{U}}\mathbf{R}^*$\\
			\quad  $\ddot{\mathbf{U}}(i,l) =  \langle l= \arg \max_{k \in [K]} \tilde{\mathbf{U}}(i,k) \rangle, i \in[n], l \in [K]$
			\STATE  find the optimal orthonormal matrix $\mathbf{R}^*$ by:\\
			\quad $\ddot{\mathbf{U}}^T\hat{\mathbf{U}} = \mathbf{V}\Omega \tilde{\mathbf{V}}^T, \quad \Omega = Diag(\omega)$\\
			\quad $\bar{\phi}  = tr(\Omega)$\\
			\quad If $|\bar{\phi} - \bar{\phi}^*| < $ machine precision, then stop and output $\ddot{\mathbf{U}}$\\
			\quad $\bar{\phi}^* = \bar{\phi}$ \\
			\quad $\mathbf{R}^* = \tilde{\mathbf{V}}\mathbf{V}^T$
			\STATE Go to step 6.
		\end{algorithmic}
	\end{algorithm}
	
	\section{Numerical Experiments}
	We have numerical experiments on the two proposed algorithms. 
	\subsection{Toy Dataset}
	\begin{figure}[H]
		\centering
		\subfigure{
			\includegraphics[width=14cm]{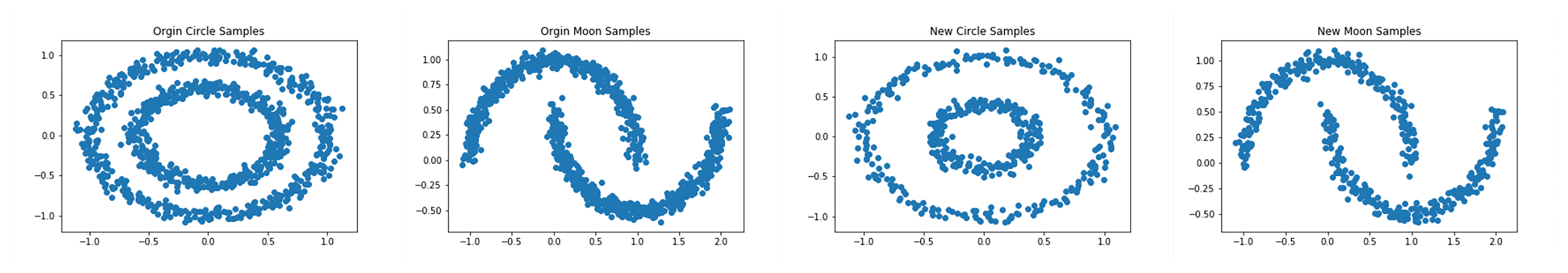}
		}
		\subfigure{
			\includegraphics[width=14cm]{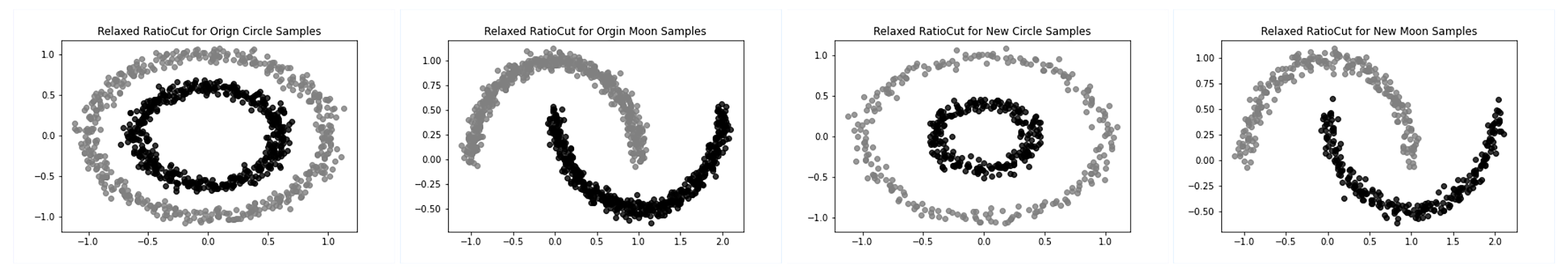}
		}
		\subfigure{
			\includegraphics[width=14cm]{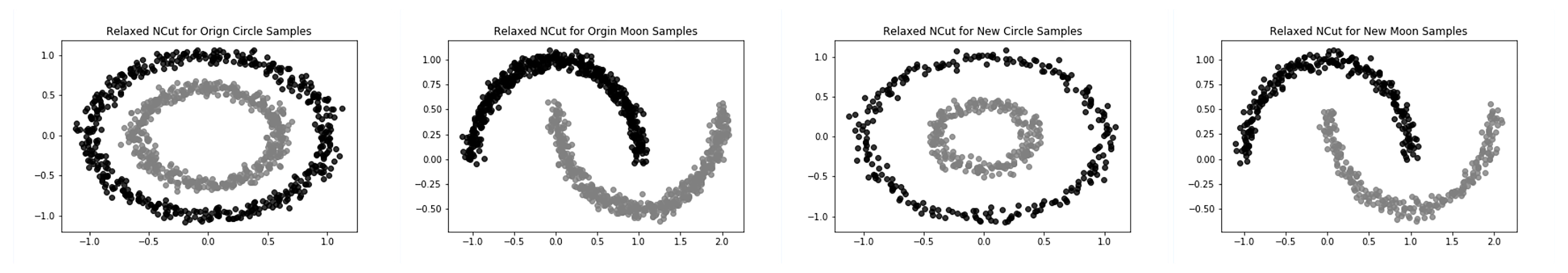}
		}
		\caption{Numerical experimental Results}
	\end{figure}
	We first verify the effectiveness of the proposed algorithms on two popular toy datasets, circle datasets and moon datasets, implemented by \emph{scikit-learn} which is a well-known tool for predictive data analysis in machine learning. The number of original samples is set as $n = 2000$, and the number of unseen samples is set as $m = 500$. The weight function $W(x,y)$ is used by Gaussian kernel function $\exp \{-\frac{\| x-y \|^2}{2\sigma^2}\}$, where $\sigma$ is set as $0.1$ for \emph{relaxed} NCut and $0.0006$ for \emph{relaxed} RatioCut. The first row is four illustrations of the data points. Among them, the first two illustrations are the original samples and denote the circle datasets and moon datasets, respectively, while the last two illustrations are out-of-sample data points. We use the eigenvectors of the original data to cluster the unseen samples without requiring the eigendecomposition on the overall samples. Specifically, we use the information in the first illustration to predict the classification label of the samples in the third illustration, and the second illustration corresponds to the fourth illustration. The second row and the third row are clustering results for \emph{relaxed} RatioCut and \emph{relaxed} NCut, respectively. Among them, the first two illustrations are spectral clustering on the original samples, while the last two illustrations are spectral clustering on the out-of-sample samples. In each illustration, the samples are assigned one color: black or gray. From row 2, one can see that the unseen data points are correctly colored and correctly classified, suggesting that our proposed algorithms can use the eigenvectors of the original samples to correctly cluster the unseen samples. Similar results hold for row 3 which corresponds to \emph{relaxed} NCut. In conclusion, from Figure 1, one can see that our two proposed algorithms are effective in clustering the unseen data points.
	\subsection{Real Dataset}
\begin{figure}[H]
		\centering
		\includegraphics[width=14cm]{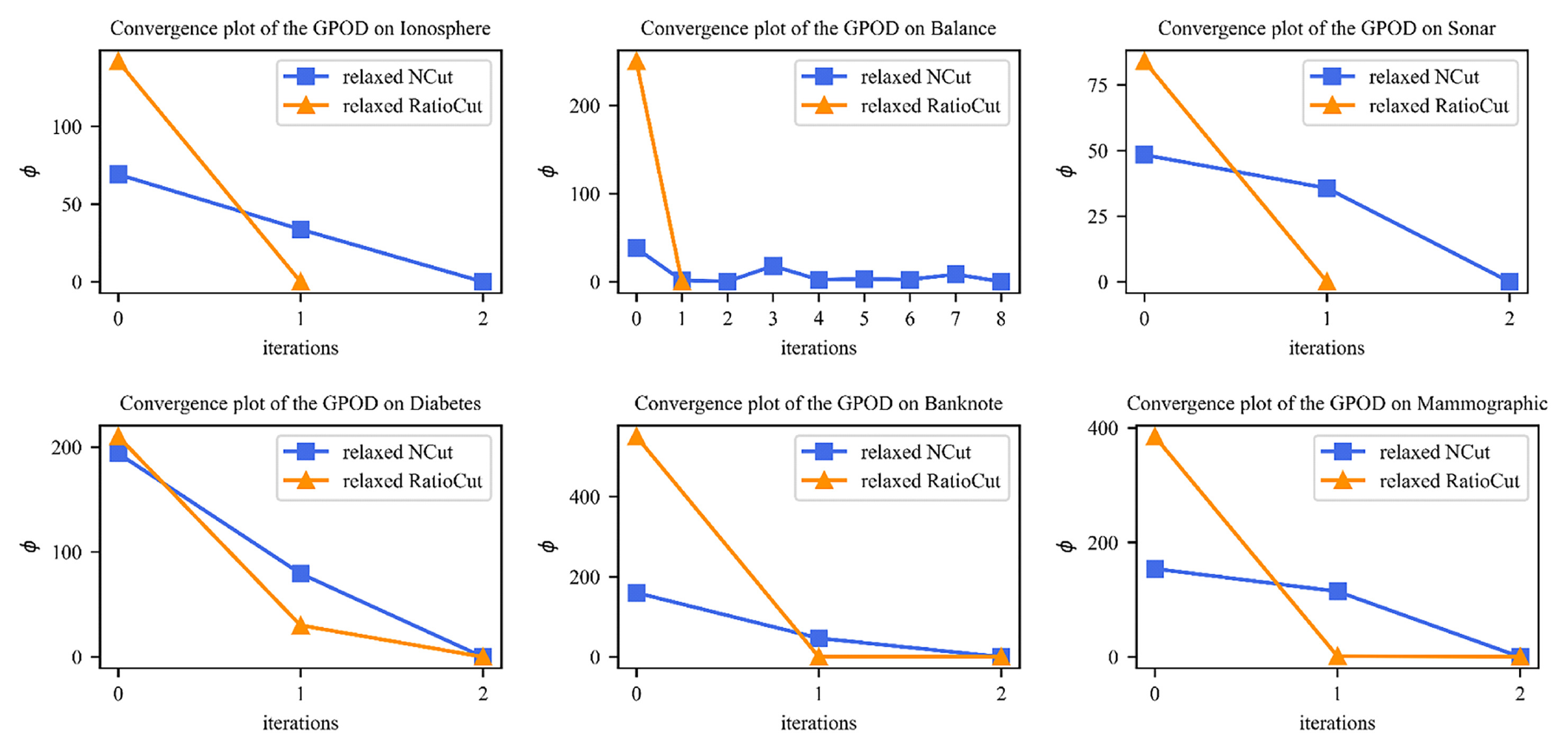}
		\caption{Convergence of the GPOD on each dataset}
		\label{fig:convergencecombin}
	\end{figure}

	Additionally, six real datasets collected from the UCI machine learning repository are used for the experiments, which are commonly used in clustering. We compare $\mathrm{GPOD}$ with the relevant algorithm $k$-means, where \textbf{$k$-means clusters the unseen data by choosing the closest cluster center}. The details of the datasets are presented in Table \ref{tab:datasets}. To measure the performance, we adopt Accuracy (ACC) and Normalized Mutual Information (NMI) as evaluation metrics. The closer the scores of these metrics are to $1$, the better performance.  For each dataset, we randomly choose $80\%$ of the samples as the training set (i.e., the original data) and the remaining $20\%$ as the testing set (i.e., the unseen samples). The Gaussian kernel function $\exp \{-\frac{\| x-y \|^2}{2\sigma^2}\}$ is chosen as the weight function $W(x,y)$ as above. There are two hyperparameters in the experiments: the number of clusters denoted as $K$ and the Gaussian kernel function parameter denoted as $\sigma$. We set $K$ equal to the number of classes in the dataset, and the settings of parameter $\sigma$ are given in Table \ref{tab:sigma}. Here the parameter settings of \emph{relaxed} NCut and \emph{relaxed} RatioCut are denoted as setting[1] and setting[2] respectively.  All algorithms are performed four times on each dataset to reduce the impact of randomness, and then the average performance is computed. From the experimental results in Table \ref{tab:result}, one can see that $\mathrm{GPOD}$ outperforms $k$-means. Meanwhile, the convergence of the $\mathrm{GPOD}$ is shown in Figure \ref{fig:convergencecombin}, where the six illustrations show the convergence speed of the algorithm on the six datasets, respectively. For each illustration, the horizontal axis represents the iteration steps and the vertical axis represents the optimization objective of the $\mathrm{POD}$ algorithm, the gap between the discrete and continuous solutions. As can be seen from Figure \ref{fig:convergencecombin}, the algorithm can converge quickly after a few iterations.
	\begin{table}[H]
		\centering
		\begin{tabular}{@{}cccc@{}}
			\toprule
			Datasets                          & Instances                 & Attributes              & Classes \\ \midrule
			\multicolumn{1}{c|}{Ionosphere}   & \multicolumn{1}{c|}{351}  & \multicolumn{1}{c|}{34} & 2       \\
			\multicolumn{1}{c|}{Balance}      & \multicolumn{1}{c|}{625}  & \multicolumn{1}{c|}{4}  & 3       \\
			\multicolumn{1}{c|}{Sonar}        & \multicolumn{1}{c|}{208}  & \multicolumn{1}{c|}{60} & 2       \\
			\multicolumn{1}{c|}{Diabetes}     & \multicolumn{1}{c|}{768}  & \multicolumn{1}{c|}{20} & 2       \\
			\multicolumn{1}{c|}{Banknote}     & \multicolumn{1}{c|}{1372} & \multicolumn{1}{c|}{5}  & 2       \\
			\multicolumn{1}{c|}{Mammographic} & \multicolumn{1}{c|}{961}  & \multicolumn{1}{c|}{6}  & 2       \\ \bottomrule
		\end{tabular}
		\caption{Characteristics of the datasets}
		\label{tab:datasets}
	\end{table}
	
	\begin{table}[H]
		\centering
		\begin{tabular}{@{}lcccccc@{}}
			\toprule
			& \multicolumn{1}{l}{Ionosphere} & \multicolumn{1}{l}{Balance} & \multicolumn{1}{l}{Sonar} & \multicolumn{1}{l}{Diabetes} & \multicolumn{1}{l}{Banknote} & \multicolumn{1}{l}{Mammographic} \\ \midrule
			\emph{relaxed} NCut     & 0.3970                         & 0.0940                      & 0.0245                    & 0.1730                       & 0.0520                       & 0.0820                           \\
			\emph{relaxed} RatioCut & 0.3970                         & 0.0940                      & 0.0245                    & 5.5000                       & 0.0480                       & 0.0820                           \\ \bottomrule
		\end{tabular}
		\caption{Settings of parameter $\sigma$}
		\label{tab:sigma}
	\end{table}
	
	\begin{table}[H]
		\centering
		\begin{tabular}{@{}cccccccc@{}}
			\toprule
			\multirow{2}{*}{Method}                       & \multirow{2}{*}{Metric}  & \multicolumn{6}{c}{Datasets}                                                                              \\ \cmidrule(l){3-8} 
			&                          & Ionosphere      & Balance         & Sonar           & Diabetes        & Banknote        & Mammographic    \\ \midrule
			\multicolumn{1}{c|}{\multirow{2}{*}{$k$-means}}  & \multicolumn{1}{c|}{ACC} & 76.1            & 52.5            & 55.7            & 67.5            & 56.0            & 64.8            \\
			\multicolumn{1}{c|}{}                         & \multicolumn{1}{c|}{NMI} & 0.1965          & 0.1399          & 0.0474          & 0.0581          & 0.0123          & 0.0710          \\ \midrule
			\multicolumn{1}{c|}{\multirow{2}{*}{GPOD[1]}} & \multicolumn{1}{c|}{ACC} & 78.2            & 64.4            & \textbf{64.8}   & 71.5            & 73.0            & \textbf{76.0}   \\
			\multicolumn{1}{c|}{}                         & \multicolumn{1}{c|}{NMI} & 0.3058          & 0.2370          & \textbf{0.0697} & 0.0978          & 0.2750          & \textbf{0.2161} \\ \midrule
			\multicolumn{1}{c|}{\multirow{2}{*}{GPOD[2]}} & \multicolumn{1}{c|}{ACC} & \textbf{79.1}   & \textbf{66.9}   & 58.2            & \textbf{74.0}   & \textbf{88.3}   & 74.8            \\
			\multicolumn{1}{c|}{}                         & \multicolumn{1}{c|}{NMI} & \textbf{0.3229} & \textbf{0.2937} & 0.0427          & \textbf{0.1417} & \textbf{0.4986} & 0.2012          \\ \bottomrule
		\end{tabular}
		\caption{Experimental results of the performance of the three methods. The best results are highlighted in bold font.}
		\label{tab:result}
	\end{table}
	
	\section{Table of Notation}
	Please refer to Table \ref{tab:notation}.
	\begin{table}[htb]
		\centering
		\begin{tabular}{@{}ccc@{}}
			\toprule
			Notation              & Description                                     & Section \\ \midrule
			$\mathcal{X}$         & a subset of $\mathbb{R}^d$                      & 2       \\
			$\rho$                & a probability measure on $\mathcal{X}$          & 2       \\
			$\rho_n$              & the empirical measure on $\mathcal{X}$          & 2       \\
			$\mathbf{X}$           & a set of samples                                & 2       \\
			$\mathcal{G} = (\mathbb{V}, \mathbb{E}, \mathbf{W})$ & weighted graph constructed on $\mathbf{X}$                 & 2   \\
			$\mathbb{V}$, $\mathbb{E}$                           & set of all nodes and edges respectively                    & 2   \\
			$W(x,y)$              & weight function                                 & 2       \\
			$\mathbf{W}$                                         & weight matrix calculated by the weight function $W(x,y)$   & 2   \\
			$|\mathbb{V}| = n $   & number of elements in set $\mathbb{V}$          & 2       \\
			$K$                   & the clustering number                           & 2       \\
			$\mathbf{D}$          & degree matrix                                   & 2       \\
			$\mathbf{L}$          & unnormalized graph Laplacian                    & 2       \\
			$\mathbf{L}_{rw}$     & asymmetric normalized graph Laplacian           & 2       \\
			$\mathbf{U}$          & a set of $K$ vectors & 2       \\
			$\hat{F}(\mathbf{U})$ & the empirical error                             & 2       \\
			$\mathrm{vol}(\mathbb{V}_j)$                         & the summing weights of edges of a subset $\mathbb{V}_j$ of a graph      & 2   \\
			$L^2(\mathcal{X},\rho)$                              & the space of square integrable functions                   & 3   \\
			$L(x,y)$              & a kernel function                               & 3.1     \\
			$\mathcal{H}$         & a reproducing kernel Hilbert space              & 3.1     \\
			$\kappa$              & the supremum $\sup_{x\in\mathcal{X}}L(x,x)$                        & 3.1     \\
			$L_K$                 & an integral operator                     & 3.1     \\
			$U$          & a set of $K$ functions & 2       \\
			$F(U)$                & the population-level error                      & 3.1     \\
			$\widetilde{U}^*$                                    &optimal solution of the minimal population-level error of \emph{relaxed} RatioCut (or NCut)& 3.1 (or 3.2) \\
			$\widetilde{\mathbf{U}}^*$                           & optimal solution of the minimal empirical error  of \emph{relaxed} RatioCut (or NCut)& 3.1 (or 3.2) \\
			$T_n$                 & a empirical operator of \emph{relaxed} RatioCut  & 3.1     \\
			$\ddot{U}$            & the population-level discrete solution          & 3.1     \\
			$\check{U}$           & consisting of the first $K$ eigenfunctions of the operator $T_n$ (or $\mathbb{L}_n$)    & 3.1 (or 3.2)     \\
			$U^*$           & the optimal solution of the minimal population-level error of RatioCut (or NCut)  & 3.1 (or 3.2)     \\
			$\mathbb{L}$          & an integral operator                              & 3.2     \\
			$\mathcal{K}$         & a continuous real-valued bounded kernel         & 3.2     \\
			$\mathbb{L}_n$, $A_n$        & empirical operators of \emph{relaxed} NCut      & 3.2     \\
			$\mathbf{R}$          & an orthonormal matrix                           & 4       \\
			$\bar{\mathbf{X}}$    & a set of new samples                            & 4.1     \\
			$\bar{\mathbf{U}}$    & the eigenvectors of the  new samples    $\bar{\mathbf{X}}$             & 4.1     \\ \bottomrule
		\end{tabular}
		\caption{Notations.}
		\label{tab:notation}
	\end{table}

\end{document}